\documentclass[smallcondensed]{svjour3}
\RequirePackage{fix-cm}

\usepackage{graphicx}
\usepackage{parskip}
\usepackage[table]{xcolor}
\usepackage{hyphenat}
\usepackage{tabularx}
\usepackage{amsfonts,amsmath,amssymb}
\usepackage{tikz}
\usepackage{hhline}
\usepackage{appendix}
\usepackage{arydshln}
\usepackage{url}

\usepackage{bbm,enumerate,subcaption}
\usepackage[linesnumbered,ruled]{algorithm2e}
\SetKwFor{For}{for (}{) $\lbrace$}{$\rbrace$}
\let\oldnl\nl% Store \nl in \oldnl
\newcommand{\nonl}{\renewcommand{\nl}{\let\nl\oldnl}}% Remove line number for one line
\SetAlgoLined\DontPrintSemicolon

\spnewtheorem*{theorem*}{Theorem}{\bfseries}{\itshape}

\newcommand{\intd}{\,\mathrm{d}}
\DeclareMathOperator*{\argmin}{arg\,min}

\begin{document}

\title{Valid prediction intervals for regression problems}

\author{Nicolas Dewolf \and Bernard De Baets \and Willem Waegeman}
\institute{
    Nicolas Dewolf \at KERMIT, Department of Data Analysis and Mathematical Modelling\\
    Ghent University, Belgium\\
    \email{nicolas.dewolf@hotmail.com}
\and
    Bernard De Baets \at KERMIT, Department of Data Analysis and Mathematical Modelling \\
    Ghent University, Belgium\\
    \email{bernard.debaets@ugent.be}
\and
    Willem Waegeman \at KERMIT, Department of Data Analysis and Mathematical Modelling\\
    Ghent University, Belgium\\
    \email{willem.waegeman@ugent.be}
}

\date{}

\maketitle

\begin{abstract}
    Over the last few decades, various methods have been proposed for estimating prediction intervals in regression settings, including Bayesian methods, ensemble methods, direct interval estimation methods and conformal prediction methods. An important issue is the validity and calibration of these methods: the generated prediction intervals should have a predefined coverage level, without being overly conservative. So far, no
    study has analysed this issue whilst simultaneously considering these four classes of methods. In this independent comparative study, we review the above four classes of methods from a conceptual and experimental point of view in the i.i.d. setting. Results on benchmark data sets from various domains highlight large fluctuations in performance from one data set to another. These observations can be attributed to the violation of certain assumptions that are inherent to some classes of methods. We illustrate how conformal prediction can be used as a general calibration procedure for methods that deliver poor results without a calibration step.
\end{abstract}

\textbf{Keywords} \,Prediction interval - calibration - regression - Bayesian network - ensemble theory - conformal prediction

\textbf{Declarations}\\
\textbf{Funding:} This research received funding from the Flemish Government under the ``Onderzoeksprogramma Artificiële Intelligentie (AI) Vlaanderen'' programme.\\
\textbf{Conflicts of interest:} None\\
\textbf{Availability of data and material:} All data sets are freely available (references are provided in the text).

\section{Introduction}

    Machine learning methods, and in particular deep learning methods, often serve as work horses in artificial intelligence systems that have a strong impact on the daily life of humans, such as self-driving cars~\cite{chen2017end,michelmore2018evaluating}, machine translation~\cite{ott2018analyzing,singh2017machine} and medical diagnostics \cite{fang2018quantile,jiang2012calibrating}. However, such systems are only accepted by humans if they exhibit a sufficient degree of reliability. As a result, analyzing what systems ``know'' and what they ``don't know'' has become an important topic of recent deep learning research, using ``uncertainty quantification'' and ``uncertainty estimation'' as prominent buzz words~\cite{NEURIPS2019_1cc8a8ea,malinin2018predictive,teye2018bayesian,van2020uncertainty}. Driven by popular application domains like computer vision and natural language processing, most of the recent literature heavily focuses on classification problems -- see e.g.~\cite{guo2017calibration,hein2019relu,naeini2015obtaining,van2020uncertainty}. Regression problems are often ignored, or at least less analyzed in such papers.

    Methods for uncertainty quantification in classification and regression problems usually differ substantially. Many traditional classification methods produce probability estimates, which are used as a starting point for uncertainty quantification, out-of-distribution detection and open-set recognition~\cite{9040673}. In regression, most traditional methods are so-called point predictors; they only predict one summary statistic of the conditional distribution (in many cases the conditional mean). However, a point predictor cannot express how confident it is of a prediction, and typically a prediction interval is returned by more complicated methods to quantify uncertainty, i.e. the wider the interval, the larger the uncertainty. Such prediction intervals can be obtained by modelling the conditional distribution in an exact or approximate manner, using for example Bayesian methods~\cite{goan2020bayesian,williams1996gaussian} or ensemble methods~\cite{kendallgal,oob_rf}. Prediction intervals can also be estimated in a more direct manner, without modelling the conditional distribution entirely, using for example quantile regression~\cite{koenker2001quantile} or conformal prediction methods~\cite{romano2019conformalized,cp_all}.

    The estimation of prediction intervals for regression has received little attention recently, and the last general review predates the ongoing deep learning wave~\cite{khosravi2011comprehensive} (at the time of writing another review appeared with a strong focus on fuzzy methods~\cite{cartagena2021review}). By now, many of the older methods have been superseded or have become unwieldy due to the complexity of modern data sets. Moreover, recent advances in deep learning have led to the introduction of some novel frameworks, such as Bayesian neural networks~\cite{goan2020bayesian}, which are typically only applied to classification problems, while having also potential for regression problems. As a result, an up-to-date comparison of methods that generate prediction intervals is necessary. This paper intends to bridge that gap, by focusing on the following aspects:

    \begin{enumerate}[i)]
        \item We give an overview of four general classes of methods that produce prediction intervals: Bayesian methods, ensemble methods, direct estimation methods and conformal prediction methods. We discuss some representative examples of each class, while paying attention to potential advantages or disadvantages.
        \item We intend to elaborate on the calibration (or validity) of prediction intervals and relate it to data and model properties. For many applications it is indeed important to have well-calibrated prediction intervals (see below for a formal definition of calibration).
        \item We show how \textit{conformal prediction} can be applied as a general framework to obtain well-calibrated prediction intervals, starting from simple point predictors and methods that generate uncalibrated intervals.
        \item We provide an in-depth experimental comparison of the four main classes of methods based on their performance across a wide range of data sets. We interpret the observed differences and discuss practical difficulties such as hyperparameter tuning and model selection based on prior knowledge.
    \end{enumerate}
    The general structure of the paper is as follows. In Section \ref{section:problem_statement} some general aspects of the estimation of prediction intervals for regression are discussed. Subsequently, in Section \ref{section:methods}, the different classes of methods are reviewed. The setup of an experimental assessment for a selection of methods is presented in Section \ref{section:methodology}. A discussion of the main results of those experiments can be found in Section \ref{section:results}.

\section{Formal problem statement}\label{section:problem_statement}

    Before being able to formally state the problem of confidence estimation, a small note on the notations used in the remainder of the text is required. In this study, only regression problems involving real-valued functions $f:\mathcal{X}\rightarrow\mathbb{R}$ will be considered, where the feature space $\mathcal{X}$ can contain both continuous and discrete ``dimensions''. Generic models and model parameters will be denoted by $\hat{y}$ and $\theta$, respectively, while a generic interval estimator (to be introduced below) will be denoted by $\Gamma$. If emphasis needs to be placed on the quantity that is estimated, the estimator will be indicated by a caret, e.g.\ an estimator for the standard deviation $\sigma$ will be denoted by $\hat{\sigma}$. The fixed confidence level will always be denoted by $\alpha$. The Cartesian product of the feature and target spaces will be denoted by $\mathcal{Z}:=\mathcal{X}\times\mathbb{R}$. A generic element of the instance space $\mathcal{Z}$ will be denoted by $\mathbf{z}\equiv(\mathbf{x},y)$, while a collection of such elements will be denoted by $\mathbf{Z}\equiv(\mathbf{X},\mathbf{y})$, where~$\mathbf{X}$ denotes the design matrix obtained from stacking the feature arrays from the individual observations. Here, the common convention is used that tuples are indicated by boldface and matrices are written in boldface capitals. A new, previously unseen instance will be denoted by an asterisk: $\mathbf{z}^*\equiv(\mathbf{x}^*,y^*)$. General data sets will be denoted by $\mathcal{D}$. These can be divided into a training, validation (or calibration) and test set, which are denoted by $\mathcal{T},\mathcal{V}$ and $\mathcal{D}^*$ respectively.

    Given a tuple of features $\mathbf{x}\in\mathcal{X}$, it is not only interesting to obtain a point prediction $\hat{y}(\mathbf{x})\in\mathbb{R}$, but often one is also interested in a region $\Gamma(\mathbf{x})\subseteq\mathbb{R}$ such that the true response $y\in\mathbb{R}$ will be contained in $\Gamma(\mathbf{x})$ with a predetermined probability. Functions $\Gamma$ that assigns such a region to every point $\mathbf{x}\in\mathcal{X}$ are generally called \textit{confidence estimators}. In this study, only interval estimators will be considered, i.e.\ functions $\Gamma:\mathcal{X}\rightarrow[\mathbb{R}]$, where $[\mathbb{R}]$ denotes the set of all closed subintervals of~$\mathbb{R}$. When constructing prediction intervals one has to specify the probability with which the estimator is allowed to make errors, since in general no realistic model, i.e.\ a model with a finite number of parameters inferred from a finite number of data points, can be perfect. Defining the \textit{coverage (probability)} of the interval estimator $\Gamma$ as:
    \begin{gather}
        \label{coverage}
        \mathcal{C}(\Gamma, P) := \mathrm{E}_{(\mathbf{x},y) \sim P}\big[\mathbbm{1}(y\in \Gamma(\mathbf{x}))\big] = \text{Prob}\{y\in\Gamma(\mathbf{x})\}\,,
    \end{gather}
    where $\mathbbm{1}$ denotes the indicator function and $P$ denotes the joint distribution on $\mathcal{Z}$, a \textit{significance} or \textit{confidence level} $\alpha$ is chosen \cite{faulkenberry1973method,fraser1956tolerance} such that
    \begin{gather}
        \label{calibrated}
        \mathcal{C}(\Gamma,P)\geq 1-\alpha
    \end{gather}
    should hold. In some cases this condition will, however, not be satisfied. To distinguish between these situations, estimators satisfying the inequality are said to be \textit{(conservatively) valid} \cite{cp_all} or \textit{calibrated} (in accordance with the classification literature \cite{guo2017calibration}). It should be clear that the best models are those that get as close as possible to saturating the inequality. Of course, one could argue that a perfect model would have a coverage of 100\%, i.e.\ it should always be completely confident, but this cannot be a realistic requirement. In real-life situations only a finite number of data samples can be provided, which implies that there always exists some uncertainty.

    Moreover, simply having an estimate of the uncertainty that satisfies condition \eqref{calibrated} is not sufficient. A trivial example would be the situation where a model always outputs the full target space as a prediction region. It clearly satisfies the validity condition~\eqref{calibrated}, but it is hard to extract any meaning from the result. The simplest measure of meaningfulness will, therefore, be the average length of the prediction intervals:
    \begin{gather}
        \label{expected_width}
        \mathcal{W}(\Gamma,P) := \mathrm{E}_{(\mathbf{x},y) \sim P}\big[|u(\mathbf{x}) - l(\mathbf{x})|\big]\,,
    \end{gather}
    where the functions $u,l$ denote the upper and lower bound of the prediction interval produced by $\Gamma$. Depending on the context (this includes both the type of data and the class of models) other, more informative, measures can be considered. However, since this study aims to be as high-level as possible, only this particular choice will be considered.

\section{Overview of methods}\label{section:methods}

    In this section, four general classes of models are discussed. The first class has its roots in probability theory and, therefore, can be expected to have better theoretical guarantees for the validity and behaviour. The second class consists of methods that are built from a collection of estimators and generally have a superior predictive performance when compared to individual models. The third class contains the models that are specifically trained to yield a prediction interval, while the last class constitutes a framework that allows to turn any given point predictor into a valid interval estimator. Many different methods are nowadays being used to produce uncertainty estimates. However, to our opinion, all methods that construct prediction intervals in a regression setting can be assigned to one of these four classes.

\subsection{Bayesian methods}\label{section:exact}

    In Bayesian inference one tries to model the distribution of interest by updating a prior estimate using a collection of observed data. The conditional distribution $p(Y\,|\,X,\mathcal{D})$ is inferred from a given parametric model or \textit{likelihood} function $p(Y\,|\,X,\theta)$, a prior distribution $p(\theta)$ over the model parameters and a data set $\mathcal{D}\equiv(\mathbf{X},\mathbf{y})$. The first step is to update the prior belief based on the data set using Bayes' rule:
    \begin{gather}
        \label{inference}
        p(\theta\,|\,\mathcal{D}) = \frac{p(\mathbf{y},\theta\,|\,\mathbf{X})}{p(\mathbf{y}\,|\,\mathbf{X})} = \frac{p(\mathbf{y}\,|\,\mathbf{X},\theta)p(\theta)}{\int p(\mathbf{y}\,|\,\mathbf{X}, \theta)p(\theta)\intd\theta}\,.
    \end{gather}
    After this posterior distribution is computed, the posterior predictive distribution is calculated by marginalizing over the parameters:
    \begin{gather}
        \label{predictive}
        p(y^*\,|\,\mathbf{x}^*,\mathcal{D}) = \int p(y^*\,|\,\mathbf{x}^*,\theta)p(\theta\,|\,\mathcal{D})\intd\theta\,.
    \end{gather}
    This process is summarized in Algorithm~\ref{algorithm:bayes}. Note that the algorithm can simply be repeated in an on-line fashion when more data becomes available. One simply takes the ``old'' posterior distribution $p(\theta\,|\,\mathcal{D})$ as the new prior distribution.

    To obtain a point estimate for future predictions, the most popular choice is the conditional mean $\mathrm{E}[y^*\,|\,\mathbf{x}^*, \mathcal{D}]$. For Bayesian methods the construction of a prediction interval reduces to the classical, but not always straightforward, problem of finding prediction intervals for a given distribution. One of the main advantages of Bayesian methods compared to other approaches is that one can incorporate domain knowledge into the prior distribution. By leveraging this knowledge to more accurately characterize the class of distributions underlying the data generating process, one can obtain improved uncertainty estimates, especially in settings with very few data.

    \begin{algorithm}[t!]
        \SetKwInOut{Input}{Input}
        \SetKwInOut{Output}{Output}

        \nonl\underline{framework Bayes} $(\mathcal{A}, \mathcal{L}, p, \mathcal{D})$\;
        \Input{Model architecture $\mathcal{A}$, likelihood function $\mathcal{L}(Y\,|\,X,\theta)$, prior distribution $p(\theta)$, data set~$\mathcal{D}$}
        \Output{Predictive distribution $p(Y\,|\,X,\mathcal{D})$\newline}
        Construct a model with architecture $\mathcal{A}$, where the parameters are sampled from $p(\theta)$\;
        Update the prior distribution through Bayes' rule~\eqref{inference}\;
        Infer the predictive distribution $p(Y\,|\, X,\mathcal{D})$ using Eq.~\eqref{predictive}\;
        \KwRet $p(Y\,|\, X,\mathcal{D})$
        \caption{Bayesian modelling}
        \label{algorithm:bayes}
    \end{algorithm}

\subsubsection*{Gaussian processes}

    One of the most popular probabilistic models for regression problems is the Gaussian process~\cite{williams1996gaussian}. The main reason for its popularity is that it is one of the only Bayesian methods where the inference step \eqref{inference} can be performed exactly, since the marginalization of multivariate normal distributions can be written in closed form and, hence, no approximations are needed (in principle). Formally, a Gaussian process (GP) is defined as a stochastic process
    for which the joint distribution of any finite number of random variables is (multivariate) Gaussian. Gaussian processes are, therefore, characterized by only two functions: the mean and covariance functions $m:\mathcal{X}\rightarrow\mathbb{R}$ and $k:\mathcal{X}\times\mathcal{X}\rightarrow\mathbb{R}$.

    If one assumes that the population is distributed according to a Gaussian process with given mean and covariance functions, by definition, the responses for the training and test instances $\mathbf{X},\mathbf{x}^*$ are distributed according to a multivariate Gaussian as follows (functions are calculated elementwise on tuples and matrices):
    \begin{gather}
        \begin{pmatrix}
            \mathbf{y}\\
            y^*
        \end{pmatrix} \sim
        \mathcal{N}\left(
        \begin{pmatrix}
            m(\mathbf{X})\\
            m(\mathbf{x}^*)
        \end{pmatrix},
        \begin{pmatrix}
            k(\mathbf{X},\mathbf{X})&k(\mathbf{X},\mathbf{x}^*)\\
            k(\mathbf{x}^*,\mathbf{X})&k(\mathbf{x}^*,\mathbf{x}^*)
        \end{pmatrix}
        \right)\equiv
        \mathcal{N}\left(
            \begin{pmatrix}
            \boldsymbol{\mu}\\
            \mu^*
        \end{pmatrix},
        \begin{pmatrix}
            \Sigma&\Sigma^*\\
            (\Sigma^*)^\mathsf{T}&\Sigma^{**}
        \end{pmatrix}
        \right)\,,
    \end{gather}
    where the notations $\boldsymbol{\mu} := m(\mathbf{X})$, $\mu^* := m(\mathbf{x}^*)$, $\Sigma := k(\mathbf{X}$, $\mathbf{X})$, etc.\ are introduced for brevity. With this convention, $\boldsymbol{\mu}$ and $\Sigma$ are, for example, a tuple and a matrix, respectively. By conditioning on the data set $\mathcal{D}$, one can obtain the posterior distribution:
    \begin{gather}
        \left.\left(y^*\,\right|\,\mathbf{x}^*,\mathcal{D}\right)\sim\mathcal{N}\Big(\mu^* + (\Sigma^*)^\mathsf{T}\Sigma^{-1}(\mathbf{y}-\boldsymbol{\mu}),\Sigma^{**} - (\Sigma^*)^\mathsf{T}\Sigma^{-1}\Sigma^*\Big)\,.
    \end{gather}
    From the latter formula it is evident that the point estimate is given by
    \begin{gather}
        \mathrm{E}[y^*\,|\,\mathbf{x}^*,\mathcal{D}] = \mu^* + (\Sigma^*)^\mathsf{T}\Sigma^{-1}(\mathbf{y} - \boldsymbol{\mu})\,.
    \end{gather}
    Although they are conceptually extremely simple, it is also immediately clear that Gaussian processes suffer from a major drawback. Due to the matrix inversion in the calculation of both the mean and covariance of the posterior distribution, they are computationally less efficient than neural network-based methods. The memory complexity of matrix inversion scales as $O\left(|\mathcal{D}|^2\right)$, while the computational complexity scales as $O\left(|\mathcal{D}|^3\right)$. For large data sets this gives rise to a considerable overhead. However, many approximations exist in the literature. Some of these try to reduce the complexity of the matrix operations by introducing stochastic approximations, while others belong to the variational class to be introduced in the next section \cite{hensman2015scalable,wilson2015kernel}.

    There is also an important remark to be made with respect to the validity condition~\eqref{calibrated}. When using GPs, two important assumptions are made. By definition it is assumed that the data is (conditionally) normally distributed and a further choice of covariance function has to be made by the user. When these assumptions do not properly characterize the data, it can be expected that the resulting distribution will also not be optimal. In general this will lead to invalid models.

\subsubsection*{Bayesian neural networks}

    Bayesian modelling can easily be implemented using neural networks. Here, the likelihood function $p(Y\,|\,X,\theta)$ in Eq.~\eqref{inference} is given by a parametric function for which the parameters are estimated by a neural network \cite{goan2020bayesian,mackay1992practical,neal}. A common example is a fixed-variance Gaussian likelihood \cite{hinton1993keeping}, where the mean is estimated by a neural network with weights $\theta$:
    \begin{gather}
        (y\,|\,\mathbf{x},\theta) \sim \mathcal{N}\big(\hat{y}^\theta(\mathbf{x}),\sigma^2\big)\,.
    \end{gather}
    Although ordinary neural networks have the benefit that even for a large number of features and weights they can be implemented very efficiently, their Bayesian incarnation suffers from a problem. The nonlinearities in the activation functions and the sheer number of parameters, although they are the features that make traditional NNs so powerful, lead to the inference steps~\eqref{inference} and \eqref{predictive} becoming intractable.

\subsubsection*{Approximate Bayesian inference}

    Both the integral in the inference step \eqref{inference} and in the prediction step \eqref{predictive} can in general not be computed exactly (conjugate priors \cite{Fink97acompendium}, such as normal distributions, form an important exception). At inference there are two general classes of approximations available:
    \begin{enumerate}
        \item Variational inference (VI):
        Instead of computing the posterior distribution through Eq.~\eqref{inference}, the problem is reformulated as a variational problem, i.e.\ the posterior distribution $p(\theta\,|\,\mathcal{D})$ is replaced by a parametric family of distributions $q(\theta;\lambda)$ and a divergence measure is optimized within this family \cite{blei2017variational}. For the Kullback-Leibler divergence $D_\text{KL}$, the loss function
        \begin{gather}
            \mathcal{L}_\text{VI}\big(\mathcal{D};\lambda) := D_\text{KL}\big(q(\theta;\lambda)\,\|\,p(\theta)\big) - \mathrm{E}\big[\log p(\mathcal{D}\,|\,\theta)\big]
        \end{gather}
        is called the \textit{variational free energy} in statistical physics terminology or \textit{evidence lower bound} (ELBO) in machine learning terminology. For Bayesian neural networks this function can be optimized using a backpropagation scheme similar to traditional neural networks~\cite{blundell2015weight}.

        \item Monte Carlo integration: The general idea of Monte Carlo (MC) integration \cite{gentle2009monte} is that for a general distribution $p$, an integral of the form
        \begin{gather}
            I[f] := \int f(x)p(x)\intd x
        \end{gather}
        is approximated by a finite sum
        \begin{gather}
            \label{mc_int}
            I[f]\approx\frac{1}{n}\sum_{i=1}^n f(x_i)\,,
        \end{gather}
        where the points $x_i$ are drawn from the distribution $p$.
    \end{enumerate}
    For making predictions, i.e.\ calculating the conditional mean, MC integration gives the following expression:
    \begin{gather}
        \mathrm{E}[y^*\,|\,\mathbf{x}^*,\mathcal{D}] = \frac{1}{n}\sum_{i=1}^n y_i\,,
    \end{gather}
    where the samples $y_i$ are drawn from the posterior distribution $p(y^*\,|\,\mathbf{x}^*,\mathcal{D})$.

\subsection{Ensemble methods}\label{section:ensembles}

    Ensemble learning is a popular approach to enhance predictions by training multiple machine learning models and aggregating the individual predictions, for example by taking the mean~\cite{krogh1996learning}. In general one can consider ensemble methods as an intermediate step between Bayesian methods
    and pure point predictors. They yield an approximation to Bayesian inference, where the trained models represent a sample in the parameter space and the aggregation is a kind of MC integration. This also implies that they have a natural notion of uncertainty. However, this uncertainty estimate does not always easily admit a probabilistic interpretation, because the ensembles are in practice often of an ad hoc nature and only roughly represent approximate Bayesian inference. The general structure of an ensemble learning scheme is summarized in Algorithm~\ref{algorithm:ensemble}.

    \begin{algorithm}[t!]
        \SetKwInOut{Input}{Input}
        \SetKwInOut{Output}{Output}

        \nonl\underline{framework Ensemble} $(R, \{(\mathcal{S}_i,\mathcal{A}_i)\}_{i=1}^R,\mathcal{D},\mathcal{E})$\;
        \Input{Number of models $R$, model architectures $\mathcal{A}_i$, sampling strategies $\mathcal{S}_i$, data set~$\mathcal{D}$, aggregation strategy $\mathcal{E}$}
        \Output{Ensemble predictor $\hat{y}$\newline}
        \For{$i = 1;\ i \leq R;\ i = i + 1$}{
            Use strategy $\mathcal{S}_i$ to obtain a data set $\mathcal{D}_i$ from $\mathcal{D}$\;
            Construct a model $\hat{y}_i$ with architecture $\mathcal{A}_i$\;
            Train model $\hat{y}_i$ on $\mathcal{D}_i$\;
        }
        Aggregate models $\{\hat{y}_i\}_{i=1}^R$ into a new model $\hat{y}$ using the strategy $\mathcal{E}$\;
        \KwRet $\hat{y}$
        \caption{Ensemble learning}
        \label{algorithm:ensemble}
    \end{algorithm}

    Every ensemble allows for a naive construction of a prediction interval \cite{heskes1997practical} when the aggregation strategy in Algorithm~\ref{algorithm:ensemble} is given by the arithmetic mean. By treating the predictions of the individual models in the ensemble as elements of a data sample, one can calculate the empirical mean and variance and use these as moment estimators for a normal distribution:
    \begin{align}
        \hat{\mu}(\mathbf{x}) &:= \frac{1}{R}\sum_{i=1}^R\hat{y}_i(\mathbf{x})\,,\\
        \hat{\sigma}^2(\mathbf{x}) &:= \frac{1}{R}\sum_{i=1}^R\big(\hat{y}_i(\mathbf{x})-\hat{\mu}(\mathbf{x})\big)^2\,,
    \end{align}
    where $R$ denotes the ensemble size. This gives rise to the following Gaussian prediction interval:
    \begin{gather}
        \label{gaussian_pi}
        \Gamma^\alpha_\text{ensemble}(\mathbf{x}^*) := \big[\hat{\mu}(\mathbf{x}^*) - z^\alpha\hat{\sigma}(\mathbf{x}^*)\,,\,\hat{\mu}(\mathbf{x}^*) + z^\alpha\hat{\sigma}(\mathbf{x}^*)\big]\,,
    \end{gather}
    where $z^\alpha$ is the two-tailed $z$-score corresponding to a significance level $\alpha$ for a standard normal distribution, e.g.\ $z^\alpha\approx 1.65$ for $\alpha=0.1$.

\subsubsection*{Random forests (and other \textit{bagged} ensembles)}

    Random forests, as introduced in \cite{breiman2001random} by Breiman, are an extension of the idea of decision trees. Their main feature is the use of \textit{bagging} \cite{breiman1996bagging}, where every decision tree in the forest is constructed from a subset of the original training set (possibly with replacement). Random forests have been shown to be robust against overfitting and, due to their nonparametric nature, they can be reliably applied in a wide variety of situations.

    The main disadvantage of random forests is the fact that they do not produce an uncertainty estimate by themselves aside from the crude construction \eqref{gaussian_pi} introduced above. There are two general methods for obtaining a more specialized prediction region from these predictors: Either one uses the \textit{out-of-bag samples} to calculate an uncertainty estimate or one uses a stand-alone procedure to extract uncertainty estimates (the former could be considered as a specific case of the latter).

    While the second approach will be discussed in Section \ref{section:cp} in more detail, the first one will be considered here. For a random forest, and in fact for any ensemble that implements the bagging procedure, it is possible to use \textit{out-of-bag} (OOB) estimates to construct prediction regions. Because these ensembles do not use the full training set to train the individual models, one can for every training sample $\mathbf{z}_i\equiv(\mathbf{x}_i,y_i)$ use the subensemble $\hat{y}_{(i)}$ of models that were not trained on $\mathbf{z}_i$ to get an estimate of the prediction accuracy. A well-established first step in this direction is the extension of the (infinitesimal) \textit{jackknife} method for variance estimation~\cite{miller1974jackknife} to bagged ensembles~\cite{efron1992jackknife,wager2014confidence}. This approach applies the general construction \eqref{gaussian_pi} to the ensemble consisting of the OOB estimators for all training samples. A related approach, first introduced in \cite{johansson2014regression} and later generalized in \cite{oob_rf}, uses the OOB estimates to construct an empirical error distribution $D:=\big\{y_i-\hat{y}_{(i)}(\mathbf{x}_i):(\mathbf{x}_i,y_i)\in\mathcal{T}\big\}$. The prediction interval for a new instance $\mathbf{x}^*$ is constructed as follows:
    \begin{gather}
        \label{OOBRF}
        \Gamma_\text{OOB}^\alpha(\mathbf{x}^*) := \left[\hat{y}(\mathbf{x}^*) + D_{[\alpha/2]}, \hat{y}(\mathbf{x}^*) + D_{[1-\alpha/2]}\right]\,,
    \end{gather}
    where $D_{[\beta]}$ denotes the empirical $\beta$-quantile of $D$. In Section \ref{section:cp} this method will be seen to reduce to a \textit{conformal predictor} in the case of symmetric error distributions as originally considered in \cite{johansson2014regression}. A similar idea was used in \cite{barber2021predictive}, where a cross-validation-based framework, called the \textit{jackknife}+, was introduced. Here, a prediction interval is obtained by using \textit{$k$-fold cross validation} or \textit{leave-one-out cross validation} to estimate the generalization error for arbitrary regression models. The above interval estimator for bagged ensembles is a specific instance of this approach. It should be noted that both the jackknife+ method and the estimator~\eqref{OOBRF}, also called the \textit{jackknife} method in \cite{barber2021predictive}, are nonparametric in contrast to the genuine jackknife estimator from \cite{wager2014confidence}.

\subsubsection*{Random forest quantile regression}

    Meinshausen \cite{meinshausen2006quantile} modified the random forest prediction method from the previous section to be able to directly estimate quantiles. Ordinary regression forests estimate the conditional mean of the response variable by taking a (weighted) average over the training instances in the leaves where the new instance belongs to. Quantile regression forests generalize this idea by using the property that the conditional distribution can be expressed as a conditional mean:
    \begin{gather}
        \label{cumulative}
        F(y^*\,|\,\mathbf{x}^*) = \mathrm{E}_{(\mathbf{x},y)\sim P}[\mathbbm{1}_{y\leq y^*}\,|\,\mathbf{x}=\mathbf{x}^*]\,,
    \end{gather}
    where $\mathbbm{1}_{y\leq y^*}$ denotes the indicator function of the set $\{y\in\mathbb{R}:y\leq y^*\}$. The algorithm starts by building an ordinary random forest and then extracts the relevant weights to estimate the conditional cumulative distribution $\hat{F}(y^*\,|\,\mathbf{x}^*)$. The (conditional) quantiles are inferred through their defining equation
    \begin{gather}
        \label{quantile}
        \hat{q}_\alpha(\mathbf{x}^*) := \inf\{y\in\mathbb{R}:\hat{F}(y\,|\,\mathbf{x}^*)\geq\alpha\}\,.
    \end{gather}
    The main benefit of this approach over ordinary quantile regressors (see Section~\ref{section:loss} below) is that one immediately obtains an approximation of the full conditional distribution and, accordingly, it is not necessary to train a separate model for each quantile. The downside of this method is its computational inefficiency. Because the full distribution needs to be modelled, the information about the leaf nodes of all trees needs to be stored and for every new data point $\mathbf{z}^*$ both the conditional distribution $\eqref{cumulative}$ and the infimum in \eqref{quantile} need to be recalculated, whereas for ordinary regression forests only the mean of every leaf node has to be stored. Some numerical libraries have introduced approximations, such as random sampling in the leaf nodes (e.g.\ the \texttt{quantregForest} package in R), but these approximations have been reported to give considerably worse predictions~\cite{quantreg}.

\subsubsection*{Dropout networks}

    Dropout layers were initially introduced as a stochastic regularization technique by Hinton et al.~\cite{srivastava2014dropout}. During each forward pass at training time, a random subset of the network weights is set to zero with probability $p$. In essence this means that a Bernoulli prior with parameter $p$ is placed over every weight. The idea behind this method was to reduce the ``co-adaptation'' behaviour of fully-connected networks, i.e., it makes it harder for the weights to work together to overfit the data. Gal and Ghahramani showed that dropout networks can be obtained as a variational approximation to deep Gaussian processes \cite{gal2016dropout}. Therefore, it can be expected that, at least when normality assumptions are satisfied, this method can outperform more ad hoc approaches.

    A second step in the development of dropout networks was the extension of the stochastic behaviour of the dropout layers to test time. By also randomly setting weights to zero during test time, an ensemble of different models can be obtained without having to retrain the model itself. Furthermore, because the Bernoulli priors induce a very sparse structure, MC integration can be expected to give a good approximation. A more in-depth study of stochastic regularization and Bayesian inference in dropout models is given in~\cite{gal2016uncertainty}.

    Kendall and Gal introduced a new method to estimate uncertainty using dropout networks~\cite{kendallgal}. Instead of having a neural network that produces a single point prediction, the network also estimates the predictive variance, thereby turning it into a \textit{mean-variance estimator}~\cite{khosravi2011comprehensive,nix1994estimating}. The loss function for this neural network contains a modified mean-squared error term and a penalty term for the variance:
    \begin{gather}
        \label{gaussian_score}
        \mathcal{L}_\text{Gauss}(\mathcal{T}) := \sum_{(\mathbf{x},y)\in\mathcal{T}}\frac{|y - \hat{y}(\mathbf{x})|^2}{2\hat{\sigma}^2(\mathbf{x})} + \frac{1}{2}\log\hat{\sigma}^2(\mathbf{x})\,.
    \end{gather}
    This loss function follows from the maximum likelihood principle with a Gaussian likelihood function $\mathcal{N}\big(\hat{y}(x),\hat{\sigma}^2(x)\big)$ or, equivalently, by assuming that the error $y-\hat{y}(x)$ is distributed according to $\mathcal{N}\big(0, \hat{\sigma}^2(x)\big)$. Instead of maximum likelihood estimation, one could also use maximum a posteriori estimation with a Gaussian or Laplace prior, so that the common $L^1$- or $L^2$-penalties can be captured.

    When making predictions, the conditional mean is again approximated by MC integration \eqref{mc_int}, i.e.\ one takes the average of multiple forward passes. The total predictive variance is given by the sum of the empirical variance of the ensemble and the variance predicted by the model itself:
    \begin{gather}
        \label{dropout_mve_var}
        \hat{\sigma}^2(\mathbf{x}) = \frac{1}{R}\sum_{i=1}^R\hat{y}_i(\mathbf{x})^2 - \left(\frac{1}{R}\sum_{i=1}^R\hat{y}_i(\mathbf{x})\right)^2 + \frac{1}{R}\sum_{i=1}^R\hat{\sigma}^2_i(\mathbf{x})\,,
    \end{gather}
    where $R$ again denotes the ensemble size. With this approach, the predictive distribution is modelled as an approximation of a uniformly-weighted Gaussian mixture model~\cite{reynolds2009gaussian}, where the parameters of every component are estimated by a model in the ensemble, by a normal distribution with the same mean and variance as the mixture model.

    The resulting prediction interval is virtually the same as \eqref{gaussian_pi}:
    \begin{gather}
        \label{mve_interval}
        \Gamma^\alpha_\mathrm{MVE}(\mathbf{x}^*) := \big[\hat{\mu}(\mathbf{x}^*) - z^\alpha\hat{\sigma}(\mathbf{x}^*)\,,\,\hat{\mu}(\mathbf{x}^*) + z^\alpha\hat{\sigma}(\mathbf{x}^*)\big].
    \end{gather}
    One can immediately expect that, analogous to general mean-variance estimators with a Gaussian prediction interval, this procedure does not give optimal intervals for data sets that do not follow a normal distribution. One of the consequences is that this model might suffer from the validity problems discussed in Section \ref{section:exact}. This issue was already highlighted in the appendix to \cite{gal2016dropout}.

\subsubsection*{Deep ensembles}

    The idea behind deep ensembles \cite{lakshminarayanan2017simple} is the same as for any ensemble technique: training multiple models to obtain a better and more robust prediction. The loss functions of most (deep) models have multiple local minima and by aggregating multiple models one hopes to take into account all these minima. From this point of view the approach by Lakshminarayanan et al.~\cite{lakshminarayanan2017simple} is very similar to that of~\cite{kendallgal}. However, the underlying philosophy is slightly different. First of all, although the same loss function is used, it is not obtained from a Bayesian framework, but rather chosen as a \textit{proper scoring rule} \cite{gneiting2007strictly}, i.e.\ a loss function for distributional forecasts for which a model can never obtain a lower loss than the true distribution (it is said to be \textit{strictly} proper if it has a unique minimum). By training a model with respect to a (strictly) proper scoring rule, the model is encouraged to approximate the true probability distribution. A disadvantage of these scoring rules is that they are only proper relative to a certain class of distributions, hence they still introduce distributional assumptions in the model. For example, the scoring rule~\eqref{gaussian_score} is only proper w.r.t.\ probability distributions with a finite second moment and strictly proper w.r.t.\ distributions that are determined by their first two moments. Secondly, instead of constructing an ensemble through MC sampling from the prior distribution, multiple models are independently trained and diversity is induced by using different initial parameters. This is motivated by a prior observation that random initialization often leads to superior performance when compared to other ensemble techniques \cite{lee2015m}.

    A further modification introduced by Lakshminarayan et al.\ is the use of adversarial perturbations. Instead of using an ordinary gradient descent step
    \begin{gather}
        \theta\longrightarrow\theta - \varepsilon\nabla_\theta\mathcal{L}(\mathbf{X},\mathbf{y};\theta)\,,
    \end{gather}
    an additional contribution is added where the training instances $\mathbf{x}$ are replaced by adversarial perturbations obtained using the Fast Gradient Sign Method \cite{FGSM} (this could be generalized by seeing the specific perturbation method as a hyperparameter of the model). For every instance $(\mathbf{x},y)$, the perturbed version is obtained as follows:
    \begin{gather}
        \mathbf{x}' := \mathbf{x} + \boldsymbol{\eta}\odot\text{sgn}\big(\nabla_{\mathbf{x}}\mathcal{L}(\mathbf{x},y;\theta)\big)\label{fgsm}\,,
    \end{gather}
    where $\odot$ denotes elementwise multiplication. This allows the constant $\boldsymbol{\eta}$ to be a vector as to accompany features with different ranges. The modified update rule is obtained by replacing the loss function in the gradient descent step with the total loss
    \begin{gather}
        \mathcal{L}_\text{tot}(\mathbf{X},\mathbf{y};\theta) := \mathcal{L}(\mathbf{X},\mathbf{y};\theta) + \mathcal{L}(\mathbf{X}',\mathbf{y};\theta)\,,
    \end{gather}
    where $\mathbf{X}'$ denotes the design matrix obtained from the perturbed features.

    Prediction intervals are constructed as in the previous section, i.e.\ a (conditionally) normal distribution is assumed and the intervals are given by Eq.~\eqref{mve_interval}. It was observed that this architecture shows improved modelling capabilities and robustness for uncertainty estimation. In~\cite{fort2019deep} the improved performance is attributed to the multimodal behaviour in model space. Fort et al.\ argue that most networks tend to have the property that they only work in one specific subspace of the model space, while deep ensembles (due to random initialization) can model different modes.

    Without the adversarial training, this model is similar to the one introduced by Khosravi et al.~\cite{khosravi2014constructing}. However, instead of training an ensemble of mean-variance estimators, an ensemble of point estimators is trained to predict $y$ and in a second step a separate estimator $\hat{\sigma}$ for the data noise is trained using loss function~\eqref{gaussian_score}, where the ensemble estimator is kept fixed.

\subsection{Direct interval estimation methods}\label{section:loss}

    The class of direct interval estimators consists of all methods that are trained to directly output a prediction interval. Instead of modelling a distribution or extracting uncertainty from an ensemble, they are trained using a loss function that is specifically tailored to the construction of prediction intervals. The general structure of this approach is summarized in Algorithm~\ref{algorithm:loss}. Because these methods are specifically made for estimating uncertainty, they can be expected to perform better than modified point estimators. However, this is also immediately their main disadvantage: they do not always produce a point estimate. Another disadvantage is that they are in general also specifically constructed for a predetermined confidence level $\alpha$. Choosing a different value for $\alpha$ requires the model to be retrained.

    \begin{algorithm}[t!]
        \SetKwInOut{Input}{Input}
        \SetKwInOut{Output}{Output}

        \nonl\underline{framework Direct} $(\mathcal{L}, \mathcal{A}, \mathcal{D})$\;
        \Input{Model architecture $\mathcal{A}$, loss function $\mathcal{L}$, data set $\mathcal{D}$}
        \Output{Interval estimator $\Gamma$\newline}
        Construct a model $\Gamma$ with architecture $\mathcal{A}$\;
        Train the model $\Gamma$ on $\mathcal{D}$ w.r.t.\ the loss function $\mathcal{L}$\;
        (Optional:) Post-process $\Gamma$ as to obtain an interval estimator\;
        \KwRet $\Gamma$
        \caption{Direct interval estimation}
        \label{algorithm:loss}
    \end{algorithm}

\subsubsection*{Quantile regression}

    Instead of estimating the conditional mean of a distribution, quantile regression models estimate the conditional quantiles. This has the benefit that one immediately obtains measures of the spread of the underlying distribution and, therefore, of the confidence of future predictions. A neural network quantile regressor $\hat{q}_\alpha$ for the $\alpha$-quantile can easily be implemented by replacing the common mean-squared error loss by the following \textit{pinball loss} or \textit{quantile loss} \cite{koenker2001quantile}:
    \begin{gather}
        \label{pinball_loss}
        \mathcal{L}_{\text{pinball}}(\mathcal{T}) := \sum_{(\mathbf{x},y)\in\mathcal{T}} \max\Big((1-\alpha)(\hat{q}_\alpha(\mathbf{x})-y), \alpha(y-\hat{q}_\alpha(\mathbf{x}))\Big)\,.
    \end{gather}
    This loss function tries to balance the number of data points below and above the (estimated) quantile. It can be obtained by reformulating the definition of quantiles as an optimization problem:
    \begin{gather}
        q_\alpha = \argmin_{q\in\mathbb{R}}\left((1-\alpha)\int_{-\infty}^q(q-y)p(y)\intd y + \alpha\int_q^{+\infty}(y-q)p(y)\intd y\right).
    \end{gather}
    By differentiating the argument on the right-hand side with respect to $q$ and equating it to~0, one obtains definition \eqref{quantile} of the $\alpha$-quantile. The pinball loss \eqref{pinball_loss} is then simply the loss function for the sample $\alpha$-quantile, i.e.\ the $\alpha$-quantile of the empirical distribution function. An intuition for this loss function can be gained from considering the example of the sample median ($\alpha=0.5$). The median $\xi$ is defined as the data point for which there are as many positive as negative residuals $r_i:=y_i-\xi$. From an optimization point of view, this definition is equivalent to that of the minimizer of the average of all absolute residuals. The pinball loss estimates other quantiles by replacing this average with a weighted average.

    To estimate multiple quantiles at the same time, one simply has to add multiple output nodes to the network and add terms of the form \eqref{pinball_loss} for every quantile. Although quantile regression models do not produce a point estimate in the form of a conditional mean, they can give a prediction of the conditional median. These models do, however, suffer from a problem of underrepresented tails of the distribution. As for every neural network, the performance becomes worse when less data is available. Because the quantiles that are relevant for interval estimation are often located in the tails of the distribution, especially for very low significance levels $\alpha$, the quantile regressors might not give optimal results.

\subsubsection*{High-Quality principle}

    As stated before, there are two quantities that are mainly used to evaluate the performance of interval estimators: the degree of coverage \eqref{coverage} and the average size of the prediction intervals~\eqref{expected_width}. The idea that optimal prediction intervals should saturate inequality \eqref{calibrated} and minimize the average size was dubbed the \textit{High-Quality} (HQ) principle by Pearce et al.\ \cite{pearce2020uncertainty,pearce2018high}. The idea to construct a loss function based on the HQ principle was first proposed by Khosravi et al.\ in \cite{khosravi2010lower}. There, the Lower-Upper Bound Estimation (LUBE) network, predicting the lower and upper bounds of the prediction interval, was trained using the following loss function:
    \begin{gather}
        \mathcal{L}_\mathrm{LUBE}(\mathcal{T}) := \frac{\mathrm{MPIW}}{r}\big(1 + \exp(\lambda\max(0, (1-\alpha) - \mathcal{C}))\big)\,,
    \end{gather}
    where $\mathcal{C}$ denotes the coverage degree \eqref{coverage} estimated on the training set $\mathcal{T}$ and, similarly, the Mean Prediction Interval Width (MPIW) is the average width \eqref{expected_width} estimated on $\mathcal{T}$:
    \begin{gather}
        \mathrm{MPIW} := \frac{1}{|\mathcal{T}|}\sum_{(\mathbf{x},y)\in\mathcal{T}}|u(\mathbf{x}) - l(\mathbf{x})|\,.
    \end{gather}
    The numbers $r$ and $\lambda$ are respectively the range of the response variable and a constant determining how much a deviation from optimal coverage should be penalized. The main idea is that the penalty should be proportional to the size of the intervals and that the penalty should be greater if the desired coverage is not obtained.

    Although intuitively sensible, this method admits no true theoretical derivation. It is derived solely from heuristic arguments. Later, Pearce et al.\ introduced an alternative to the LUBE loss where some of the ad-hoc choices were formalised~\cite{pearce2018high}. The most important modifications are the replacement of MPIW by \textit{captured} MPIW, i.e.\ only the lengths of the intervals that contain the true value are taken into account to avoid artificially shrinking already invalid intervals, and replacing the exponential function by a term derived from a likelihood principle (the number of captured points can be modelled by a binomial distribution). The resulting loss function is
    \begin{gather}
        \mathcal{L}_\mathrm{QD}(\mathcal{T}) := \mathrm{MPIW}_\mathrm{capt} + \lambda\frac{n}{\alpha(1-\alpha)}\max\big(0, (1-\alpha) - \mathcal{C}\big)^2\,,
    \end{gather}
    where $n$ is the total number of data points and $\lambda$ is a Lagrange multiplier determining the relative importance of the two contributions. Although this loss function is slightly more substantiated, it remains heuristic in nature and, hence, it does not admit any theoretical guarantees regarding its performance.

\subsection{Conformal prediction}\label{section:cp}

    In this section a framework is considered that can both turn any point predictor into an interval estimator and turn any existing interval estimator into a valid estimator. As originally introduced \cite{vapnik,saunders1999transduction,vovk1999machine}, the \textit{conformal prediction} (CP) framework allows for the construction of valid prediction regions given a data set and a \textit{nonconformity measure} that evaluates how different a new instance is from the given data set. The original implementation suffered from a high computational overhead because in general all calculations, including the training of any underlying models, had to be redone for every data point. A solution, called \textit{inductive conformal prediction} (ICP) or \textit{split-conformal prediction}, where the training phase of the underlying models was decoupled from the ``conformalization'' phase, was later introduced \cite{papadopoulos2002inductive,icp}. Although the theoretical guarantees for ICP are not as strong as for the original \textit{transductive} framework, the computational speed-up outweighs this drawback and, therefore, only ICP will be considered in this study.

    A short introduction to the inductive conformal prediction framework, along the lines of \cite{cp_all}, is in order. It will be assumed that the following data has been provided:
    \begin{enumerate}[1)]
        \item a data set $\mathcal{D}$, and
        \item a function $A:\mathcal{Z}\rightarrow\mathbb{R}$ (meeting some technical conditions).
    \end{enumerate}
    The real-valued function $A$ is called the \textit{nonconformity measure} in the conformal prediction literature (note that it is not a measure in the mathematical sense). A common example is the following one:
    \begin{gather}
        \label{dist}
        A_\text{point}(\mathbf{x},y) := |y - \hat{y}(\mathbf{x})|\,,
    \end{gather}
    where $\hat{y}$ can be any point predictor. Although it is the most frequently used type of nonconformity measure, not all nonconformity measures of relevance are of this form (see the next section).

    Given a nonconformity measure $A$, possibly depending on a training set $\mathcal{T}\subset\mathcal{D}$, the framework proceeds by constructing a calibration ``curve'' using the complement $\mathcal{V}:=\mathcal{D}\backslash\mathcal{T}$. For all $z_i\in\mathcal{V}$ the associated nonconformity score is defined as follows:
    \begin{gather}
        \label{score}
        \alpha_i := A(\mathbf{z}_i)\,.
    \end{gather}
    The interval estimator $\Gamma^\alpha$, associated with $A$ and $\mathcal{D}$, is then defined as the function that constructs a region $\Gamma^\alpha(\mathbf{x}^*)$ for all $\mathbf{x}^*\in\mathcal{X}$ consisting of all $y^*\in\mathbb{R}$ for which the nonconformity score $A(\mathbf{z}^*)$ is smaller than the empirical $\big((1-\alpha)(1 + \frac{1}{|\mathcal{V}|})\big)$-quantile of $\{\alpha_i:\mathbf{z}_i\in\mathcal{V}\}$. The general procedure for constructing $\Gamma^\alpha$ is shown in Algorithm~\ref{algorithm:icp}. The following theorem, which can be found in most of the conformal prediction literature, is the reason why this framework is so powerful.

    \begin{theorem*}[Marginal validity]
        If the nonconformity scores $\{a_i:\mathbf{z}_i\in\mathcal{V}\}\cup\{\alpha^*\}$ are exchangeable for all validation sets $\mathcal{V}$ and test points $\mathbf{z}^*$, the estimator $\Gamma^\alpha$ is (conservatively) valid:
        \begin{gather}
            \mathrm{Prob}\big\{y\in\Gamma^\alpha(\mathbf{x}\mid\mathcal{V})\big\}\geq1-\alpha\,,
        \end{gather}
        where the probability is taken over both $\mathbf{x}$ and $\mathcal{V}$. The estimator has been denoted by $\Gamma^\alpha(\,\cdot\mid\mathcal{V})$ to explicitly show the dependence of Algorithm~\ref{algorithm:icp} on the calibration set $\mathcal{V}$.
    \end{theorem*}

    In practice, one only has access to a single (calibration) data set and, consequently, one cannot take the expectation value over all possible data sets as is necessary to obtain marginal validity. However, for large calibration sets, the following theorem shows that one can expect approximately valid prediction intervals.
    \begin{theorem*}[Asymptotic validity]
        If the nonconformity scores $\{a_i:\mathbf{z}_i\in\mathcal{V}\}\cup\{\alpha^*\}$ are exchangeable for all test points $\mathbf{z}^*$, the estimator $\Gamma^\alpha$ becomes (conservatively) valid when $|\mathcal{V}|\rightarrow\infty$.
    \end{theorem*}

    \SetKwFunction{proc}{$\Gamma^\alpha$}
    \SetKwProg{myproc}{procedure}{}{}
    \SetKwBlock{DummyBlock}{}{}
    \begin{algorithm}[t!]
        \SetKwInOut{Input}{Input}
        \SetKwInOut{Output}{Output}

        \nonl\underline{function ICP} $(\alpha,\mathcal{T},\mathcal{V},A)$\;
        \Input{Significance level $\alpha$, nonconformity measure $A$, training set $\mathcal{T}$ and calibration set $\mathcal{V}$}
        \Output{Interval estimator $\Gamma^\alpha$\newline}
        (Optional) Train the underlying model of $A$ on $\mathcal{T}$\;
        \ForEach{$(\mathbf{x}_i,y_i)\in\mathcal{V}$}{
            Apply \eqref{score}: $\alpha_i\leftarrow A(\mathbf{x}_i,y_i)$\;
        }
        Determine the critical value of $A$: $\alpha^*\leftarrow\big((1-\alpha)(1 + \frac{1}{|\mathcal{V}|})\big)\text{-quantile of }\{\alpha_i:(\mathbf{x}_i,y_i)\in\mathcal{V}\}$\;
        Construct an interval estimator as follows:\;
        \nonl\SetAlgoNoLine\DummyBlock{\SetAlgoLined
            \myproc{\proc{$\mathbf{x}:\mathcal{X}$}}{
            \ForEach{$y\in\mathbb{R}$}{
                Apply \eqref{score}: $\alpha_y\leftarrow A(\mathbf{x},y)$\;
            }
            \KwRet$\{y\in\mathbb{R}:\alpha_y\leq\alpha^*\}$\;
            }
        }\nonl\;
        \KwRet$\Gamma^\alpha$
        \caption{Inductive Conformal Prediction}
        \label{algorithm:icp}
    \end{algorithm}

    The exchangeability assumption \cite{cp_all} is the reason why the training and calibration steps in Algorithm~\ref{algorithm:icp} are performed on disjoint data sets. Note that this assumption is strictly weaker than being i.i.d.\ and will in most cases be, at least approximately, satisfied.

    An important remark regarding this framework is that it is nonparametric and almost completely assumption-free. Except for the exchangeability assumption, this framework can work with any choice of data set, predictor or nonconformity measure. This not only makes it extremely versatile and applicable in almost any setting, it also assures that there can be no negative consequences coming from broken assumptions, something many other models do suffer from (see also further below in the section ``Validity issues''). However, although virtually any function can act as a nonconformity measure, the specific choice can strongly influence the size of the resulting prediction regions.

\subsubsection*{Point predictors}

    The nonconformity measures defined in \eqref{dist} are important due to the fact that they allow to turn a standard point predictor into a (valid) interval estimator. Because this choice of nonconformity measure only considers differences in the target space, i.e.\ there is no $\mathcal{X}$-dependent scaling, the algorithm produces prediction intervals of constant size. This implies that the critical value from Algorithm~\ref{algorithm:icp} needs to be calculated only once and, therefore, that the prediction interval for a new instance $\mathbf{x}^*$ can be constructed with constant time complexity as follows:
    \begin{gather}
        \Gamma^\alpha_{\text{point}}(\mathbf{x}^*) := \left[\hat{y}(\mathbf{x}^*) - \alpha^*, \hat{y}(\mathbf{x}^*) + \alpha^*\right]\,,
    \end{gather}
    where $\hat{y}$ is the given point predictor.

    The above procedure gives uniform (or homoscedastic) prediction intervals, which is in stark contrast with most bona fide interval estimators. Although computationally simple, it ought to be clear that this is not the generic situation. Different modifications to obtain heteroscedastic models have been proposed in the literature, the main one being to \textit{normalize} \cite{papadopoulos2008normalized} the nonconformity measure by a \textit{dispersion} function $\sigma:\mathcal{X}\rightarrow\mathbb{R}$. This modification is often called \textit{locally adaptive} or \textit{normalized conformal prediction}. In the case of mean-variance estimators this could be the estimate of the predictive variance (or standard deviation), while for other models a popular choice is the empirical uncertainty of a data subset found using for example a $k$-nearest neighbour algorithm \cite{papadopoulos2011regression}. It should be clear that the latter choice, although interesting because it takes into account the number of similar data points and, hence, the data uncertainty, does introduce extra computational overhead.

    Ensemble methods that make use of bagging, such as random forests, give rise to their own natural choice of both homoscedastic and heteroscedastic nonconformity measures. As mentioned in Section \ref{section:ensembles}, with bagging, every individual subestimator only uses a subset of the training set and, therefore, one can use for every training instance the subensemble for which this instance was not used during training to produce an independent prediction. This allows the full data set to act as both a training and calibration set in stark contrast to the situation of ordinary CP methods, which are restricted by the exchangeability assumption. To this end, the following modified nonconformity measure was introduced by Johansson et al.~\cite{johansson2014regression}:
    \begin{align}
        A_\text{cal}(\mathbf{x}_i,y_i) &:= |y_i-\hat{y}_{(i)}(\mathbf{x}_i)|\\
        A_\text{test}(\mathbf{x}^*,y^*) &:= |y^*-\hat{y}(\mathbf{x}^*)|\,,
    \end{align}
    where $\hat{y}_{(i)}$ denotes the OOB estimator for the data point $(\mathbf{x}_i,y_i)$. Although it cannot be formally proven that this procedure satisfies the same validity properties as ordinary ICP, the fact that OOB estimates often overestimate the uncertainty make it plausible that the above measure will lead to (conservatively) valid prediction intervals. One should also note that this approach recovers the estimate~\eqref{OOBRF} introduced in \cite{oob_rf} whenever the residuals for the calibration set $\mathcal{V}$ are distributed symmetrically around zero. To obtain formal guarantees, the authors later showed that this method can be modified \cite{bostrom2017accelerating}. Instead of using the asymmetric nonconformity measure above, they proposed the following scheme for every new observation~$\mathbf{x}^*$:
    \begin{enumerate}
        \item Choose a random training instance $\mathbf{z}_i$.
        \item Use the OOB predictor $\hat{y}_{(i)}$ to calculate the nonconformity score of a candidate $\mathbf{z}^*$ (in the process also producing a point estimate).
        \item Use the calibration scores of the complementary calibration set $\mathcal{T}\backslash\{\mathbf{z}_i\}$ to derive the prediction region for $\mathbf{x}^*$.
    \end{enumerate}
    Although this method is theoretically valid, it does require the recalculation of the critical value for every new data point, thereby introducing some extra computational overhead. Furthermore, following from the fact that the point prediction is only made using a fraction of the total forest, the predictive performance can be expected to be worse (especially for small forests).

    Similar to how the interval estimator \eqref{OOBRF} for bagged ensembles could be extended to a general cross-validation inspired framework \cite{barber2021predictive}, the above approach for conformal prediction can also be extended to a framework in which the full data set is used to calibrate the model through a $k$-fold or leave-one-out approach \cite{vovk2015cross}. For this reason they are called \textit{cross-conformal predictors}. As shown in \cite{barber2021predictive}, these methods always produce confidence regions that are contained in the intervals coming from jackknife+ estimators, with the added complexity that they are not necessarily connected, i.e. they might be disjoint unions of intervals.

\subsubsection*{Validity issues}\label{section:calibration}

    There exists a wide variety of methods that are able to handle the task of generating prediction intervals, but most of them suffer from a major problem. In practice the coverage defined in~\eqref{coverage} will not always satisfy the validity requirement~\eqref{calibrated}, which implies that the predictions of these methods are not as meaningful as desired. They can still be used as a heuristic uncertainty measure, but they lack interpretability.

    There are two main factors that contribute to this problem. First of all, as mentioned before, due to the complexity of present-day data sets, common assumptions can be violated. A simple example is the case of empirical averaging \eqref{gaussian_pi}, where the standard deviation will only give rise to calibrated intervals if the residuals follow a Gaussian distribution. Secondly, it is possible that the interval estimators are only asymptotically valid and that the size of the data set is not sufficiently large. Since the validity of typical statistical techniques is only guaranteed to hold for infinitely large data sets, there is no guarantee that these will hold for $|\mathcal{D}|\ll\infty$.

    In the context of classification problems, where especially the former issue plays a role \cite{guo2017calibration}, a wide variety of calibration methods is available: Platt scaling, temperature scaling, isotonic regression, etc.\ In general these methods take the output distribution of the trained predictor and modify it such that it becomes (approximately) well calibrated. However, these methods explicitly use the fact that the target space is finite in those cases. For regression problems, these methods can therefore not be applied in general. A convenient solution is given by the conformal prediction framework introduced above. It cannot only be used to turn a point predictor into a valid interval estimator, but it can also be used to calibrate an existing model.

\subsubsection*{Calibrating prediction intervals}

    In this section the models that predict the lower and upper bounds of prediction intervals are considered, for example the $\alpha/2$- and $(1-\alpha/2)$-quantile estimates for a given significance level $\alpha$. For this class of estimators a reasonable choice of nonconformity measure is
    \begin{gather}
        A_\text{PI}(\mathbf{x}_i,y_i) := \max\left(\hat{l}(\mathbf{x}_i)-y_i, y_i-\hat{u}(\mathbf{x}_i)\right)\,,
    \end{gather}
    as originally introduced for quantile regression by Romano et al.\ in \cite{romano2019conformalized}. As was the case for point predictors and Eq.\ \eqref{dist}, this function only considers the distance between points in the target space and, therefore, the critical value of the ICP Algorithm~\ref{algorithm:icp} will be independent of the underlying features. This again allows one to compute the critical value only once and construct a modified prediction interval for every new instance $\mathbf{x}^*$ as follows:
    \begin{gather}
        \Gamma^\alpha_\text{int}(\mathbf{x}^*) := \left[\hat{l}(\mathbf{x}^*) - \alpha^*, \hat{u}(\mathbf{x}^*) + \alpha^*\right]\,.
    \end{gather}
    The idea behind this construction is very similar to that for point predictors. One estimates the amount by which the constructed intervals are too small (or wide) on average on the calibration set and corrects for these errors in the future. Two other normalized nonconformity measures were considered in \cite{kivaranovic2020adaptive,sesia2020comparison}. However, it was empirically observed that the inherent heteroscedasticity of quantile regression models is already sufficient and that the added normalization only increases the size of the prediction intervals without gaining much in terms of validity.

    In the section on quantile regression it was noted that this approach tends to have a problem with modelling the tails of the distribution, with the added consequence that this can influence the validity at extreme significance levels. However, when combining such models with conformal prediction, validity is not an issue. Therefore, Romano et al.~\cite{romano2019conformalized} introduced a ``softening'' factor $w$ such that the underlying quantile regression model only has to estimate the $w\alpha/2$- and $(1-w\alpha/2)$-quantiles. This avoids the problems with underrepresented tails, while preserving the validity of the model (potentially excessively broadening the intervals if the smoothening parameter is not correctly tuned).

\subsection{Summary}

    In the preceding four sections, we introduced different classes of interval estimators, each having its own characteristics. In this section, we summarize the main properties for clarity and convenience. We identify four properties that are important for practical purposes. The first one is the main notion of this paper, namely validity, i.e.\ whether a model is guaranteed to produce (approximately) valid prediction intervals. Since none of the methods are valid for any finite-size data set, we only consider validity in the sense of the Marginal Validity Theorem of Section~\ref{section:cp}. The second property, which is becoming more and more important due to the increase in data size and complexity, is scalability. A third feature, which was not highlighted in the foregoing sections, but which might be important for certain applications is the possibility to include domain knowledge. Last but not least we indicate whether it is necessary to include a validation set, which is especially critical when available data is limited. The summary is shown in Table~\ref{table:method_summary}. For the four classes of methods, we also list the main references.

    \begin{table}[ht!]
        \centering
        \renewcommand{\arraystretch}{1.5}
        \begin{tabularx}{\linewidth}{|X||X|X|X|X|}
            \firsthline
            \multicolumn{1}{|c||}{Method}&\multicolumn{1}{c|}{Marginal validity}&\multicolumn{1}{c|}{Scalability}&\multicolumn{1}{c|}{Domain knowledge}&\multicolumn{1}{c|}{Validation set}\\
            \hline
            Bayesian methods\newline\cite{blei2017variational,blundell2015weight,goan2020bayesian,hensman2015scalable,williams1996gaussian,wilson2015kernel}&
            No (except for exact inference with correct priors)&
            Only scalable with approximate inference&
            Yes&
            No\\
            \hline
            Ensemble methods\newline\cite{heskes1997practical,kendallgal,lakshminarayanan2017simple,meinshausen2006quantile,nix1994estimating,wager2014confidence}&
            No&
            Yes (when scalable models are used)&
            No&
            No\\
            \hline
            Direct interval estimation\newline\cite{koenker2001quantile,pearce2020uncertainty}&
            No&
            Yes&
            No&
            Yes\\
            \hline
            Conformal prediction\newline\cite{barber2021predictive,sesia2020comparison,vovk2015cross,cp_all}&
            Yes&
            Yes (for ICP)&
            No&
            Yes\\
            \lasthline
        \end{tabularx}
        \caption{Summary of method characteristics and references.}
        \label{table:method_summary}
    \end{table}

\section{Experimental setup}\label{section:methodology}

    In this and the following section some of the models introduced above are experimentally investigated. They are evaluated and compared based on some general performance measures. Moreover, some general conclusions that can be used in future applications or research are derived.

\subsection{Quality metrics}

    An optimal interval estimator should satisfy some conditions. To assess the quality of the models, the HQ principle from Section \ref{section:loss} is adopted. First of all a model ought to be valid (or calibrated) in the sense of Eq.~\eqref{calibrated}. The more a model deviates from being well calibrated, the less reliable it becomes since the results cannot be trusted and relied upon. At the same time the results should also be interpretable, i.e.\ the end-user should be able to use the uncertainty information to make further decisions. The most obvious interpretation would be where small prediction intervals correspond to situations where the model is very confident. Therefore, the models are also compared based on the width of the prediction intervals. This is closely related to the fact that inequality \eqref{calibrated} should not only be satisfied, but should even be saturated. To this end, the following two assessment measures are used:
    \begin{enumerate}
        \item Satisfy or even saturate the validity condition \eqref{calibrated}. This is measured by approximating the coverage probability \eqref{coverage} by a finite sum over the test set:
        \begin{gather}
            \mathcal{C}(\Gamma,P) \approx \frac{1}{|\mathcal{D}^*|}\sum_{(\mathbf{x},y)\in\mathcal{D}^*}\mathbbm{1}(y\in\Gamma(\mathbf{x}))\,.
        \end{gather}
        \item Minimize the average width of the intervals, as defined in Eq.~\eqref{expected_width}:
        \begin{gather}
            \mathcal{W}(\Gamma,P) \approx \frac{1}{|\mathcal{D}^*|}\sum_{(\mathbf{x},y)\in\mathcal{D}^*}|u(\mathbf{x}) - l(\mathbf{x})|\,.
        \end{gather}
    \end{enumerate}
    Aside from the above quality measures, some other quantities might be of importance depending on the application. In the experiments only the predictive power is considered. The benefit of working with models that are built upon or include a point predictor is that one also gets a direct estimate of the response variable. Since this is important in many situations, the $R^2$-coefficients are reported (as noted in the section on quantile regression, this method cannot estimate the conditional mean, instead the conditional median is estimated).

\subsection{Data}

    Most of the data sets were obtained from the UCI repository \cite{Dua2019}. Specific references are given in Table \ref{tab:datasets}. This table also shows the number of data points and (used) features and the skewness and (Pearson) kurtosis of the response variable. All data sets were standardized (both features and target variables) before training. The data sets \texttt{blog} and \texttt{fb1} were also analysed after first taking a log transform of the response variable because these data sets are extremely skewed, which is reflected in the high skewness and kurtosis, as shown in the fourth column of Table \ref{tab:datasets}, and are believed to follow a power law distribution. This strongly improved the $R^2$-coefficient of the various models, but did not improve the prediction intervals, and therefore, these results are not included. The \texttt{crime} data set comes in two versions: the original data set consists of integer-valued data (count data), while the version used here was preprocessed using an unsupervised standardization algorithm \cite{redmond2002data}. Although standardized, the data set retains (some of) its count data properties. The \texttt{traffic} data set, aside of being very small, is also extremely sparse (on average 14 features are zero). It should be noted that all of the data sets used in this study were considered as ordinary (static) data sets. Even though some of them could be considered in a time series context, no autoregressive features were additionally extracted. The main reason to exclude autoregressive features is that most, if not all, methods considered in this study assume the data to be i.i.d. (or exchangeable), a property that is generically not valid for autoregressive data.

    \begin{table}[t!]
        \centering
        \renewcommand{\arraystretch}{1.2}
        \begin{tabular}{|c|c|c|c|c|}
            \hline
            Name&\# samples&\# features&Skewness\ /\ Kurtosis&Source\\
            \hline
            \texttt{concrete}&1030&8&0.42\ /\ 2.68&\cite{yeh1998modeling}\\
            \texttt{naval}&11934&14&0.0\ /\ 1.8&\cite{coraddu2016machine}\\
            \texttt{turbine}&9568&4&0.31\ /\ 1.95&\cite{kaya2012local}\\
            \texttt{puma32H}&8192&32&0.02\ /\ 3.04&\cite{corke1996robotics}\\
            \texttt{residential}&372&105&1.26\ /\ 5.15&\cite{rafiei2016novel}\\
            \texttt{crime2}&1994&123&1.52\ /\ 4.83&\cite{redmond2002data}\\
            \texttt{fb1}&40949&54&14.29\ /\ 301.44&\cite{singh2015comment}\\
            \texttt{blog}&52397&280&12.69\ /\ 235.3&\cite{buza2014feedback}\\
            \texttt{traffic}&135&18&1.07\ /\ 3.74&\cite{cbic2011}\\
            \texttt{star}&2161&39&0.29\ /\ 2.63&\cite{star_data}\\
            \hline
        \end{tabular}
        \caption{Overview of the data sets.}
        \label{tab:datasets}
    \end{table}

\subsection{Architectures and training}

    All neural networks were constructed using the default implementations from\linebreak \texttt{PyTorch}~\cite{pytorch}. The general architecture for all neural-network-based models was fixed. The \textit{Adam} optimizer was used for weight optimization with a fixed learning rate of $5\times10^{-4}$, in accordance with \cite{romano2019conformalized}. The number of epochs was limited to 100, unless stated otherwise. All neural networks contained only a single hidden layer with 64 neurons. The activation functions after the first layer and the hidden layer were of the ReLU type, while the activation function at the output node was simply a linear function.

    For each of the four classes of interval estimators in Section \ref{section:methods}, at least one example was chosen for a general comparison. Furthermore, to handle calibration issues, conformal prediction was chosen as a post-hoc method due to its nonparametric and versatile nature. Every model that produces a prediction interval (or an uncertainty estimate) is also re-examined after \textit{conformalization}. Hyperparameter optimization, if applicable, was performed on a validation set containing 5\% of the training samples, unless stated otherwise. The specific architectures are as follows:\footnote{All code is made publicly available at \url{https://github.com/nmdwolf/ValidPredictionIntervals}.}
    \begin{enumerate}
        \item Bayesian methods:
        \begin{enumerate}[(i)]
            \item Gaussian Process: As kernel a standard radial basis function (RBF kernel) was used. This automatically incorporates out-of-distribution uncertainty. The default implementation from the \texttt{GPyTorch} library \cite{gardner2018gpytorch} was used. This library provides a multitude of different approximations and deep learning adaptions for Gaussian processes. A heteroscedastic likelihood function with trainable noise parameter was used and the model was trained for 50 epochs (following the library's heuristics). Optimization of the noise parameter was handled internally, so no additional validation set was required.
            \item Approximate GP: Because in most present-day cases the data sets are too large to efficiently apply Gaussian processes, a variational approximation was included. The Stochastic Variational GP approximation~\cite{hensman2015scalable}, implemented in \texttt{GPyTorch}, was chosen. The same likelihood function and number of epochs were used as for the exact model.
        \end{enumerate}
        \item Ensemble methods:
        \begin{enumerate}[(i)]
            \item Dropout ensemble: An ensemble average of 50 MC samples was used. An early stopping criterion, based on the minimum of the loss function, was employed to avoid overfitting. The dropout probability was optimized over the interval $[0.05,0.5]$ with steps of $0.05$.
            \item Mean-variance estimator: The Gaussian dropout-based mean-variance estimator from \cite{kendallgal} was chosen because this model extends the dropout ensemble by explicitly incorporating uncertainty. An ensemble average of 50 MC samples was used. The dropout probability was optimized over the interval $[0.05,0.5]$ with steps of $0.05$.
            \item Deep ensembles: The ensemble consisted of 5 estimators and the adversarial step size equalled 0.01 times the range of the corresponding dimension (cf.\ \cite{lakshminarayanan2017simple}). At training time $L^2$-regularization with $\lambda=10^{-6}$ was applied.
        \end{enumerate}
        \item Direct interval estimation methods:
        \begin{enumerate}[(i)]
            \item Neural network quantile regression: The same softening factor $w=2$ as in \cite{romano2019conformalized,sesia2020comparison} was used. The early stopping criterion for neural networks was also modified to work with the average length and coverage degree of the prediction intervals instead of the loss function of the network. Dropout with $p=0.1$ and $L^2$-regularization with $\lambda=10^{-6}$ were applied at training time.
        \end{enumerate}
        \item Conformal prediction:
        \begin{enumerate}[(i)]
            \item Neural network: A standard neural network point predictor was chosen as a baseline model. Early stopping as for the ensemble methods was used as a regularization method. Furthermore, dropout with $p=0.1$ and $L^2$-regularization with $\lambda=10^{-6}$ were applied at training time.
            \item Random forest: The default implementation from \texttt{scikit-learn} \cite{scikit-learn} was used. The number of trees was fixed at 100, while all other hyperparameters were left at their default value.
        \end{enumerate}
    \end{enumerate}

    In the remainder of the text, the selected models are abbreviated as follows:

    \begin{table}[ht!]
        \centering
        \renewcommand{\arraystretch}{1.5}
        \begin{tabular}{|c|c||c|c|}
            \hline
            Model&Abbreviation&Model&Abbreviation\\
            \hline
            Neural network&NN&Random forest&RF\\
            Quantile regressor&QR&Deep ensemble&DE\\
            Dropout ensemble&Drop&Gaussian dropout-MVE&MVE\\
            Exact Gaussian Process&GP&Stochastic Variational GP&SVGP\\
            \hline
        \end{tabular}
    \end{table}
    All experimental results were obtained by evaluating the models on 50 different train/test-splits of the data sets in Table~\ref{tab:datasets}. The test set always contained 20\% of the data. If a calibration set was needed for post-hoc calibration, the training set was further divided into two equal-sized sets. Because the models were tested both with and without post-hoc calibration, this approach also allows to see how the performance changes when halving the size of the training set. The confidence level was fixed at $0.9$.

\section{Results and discussion}\label{section:results}

    \begin{figure}[p]
        \centering
        \includegraphics[width=\textwidth,keepaspectratio]{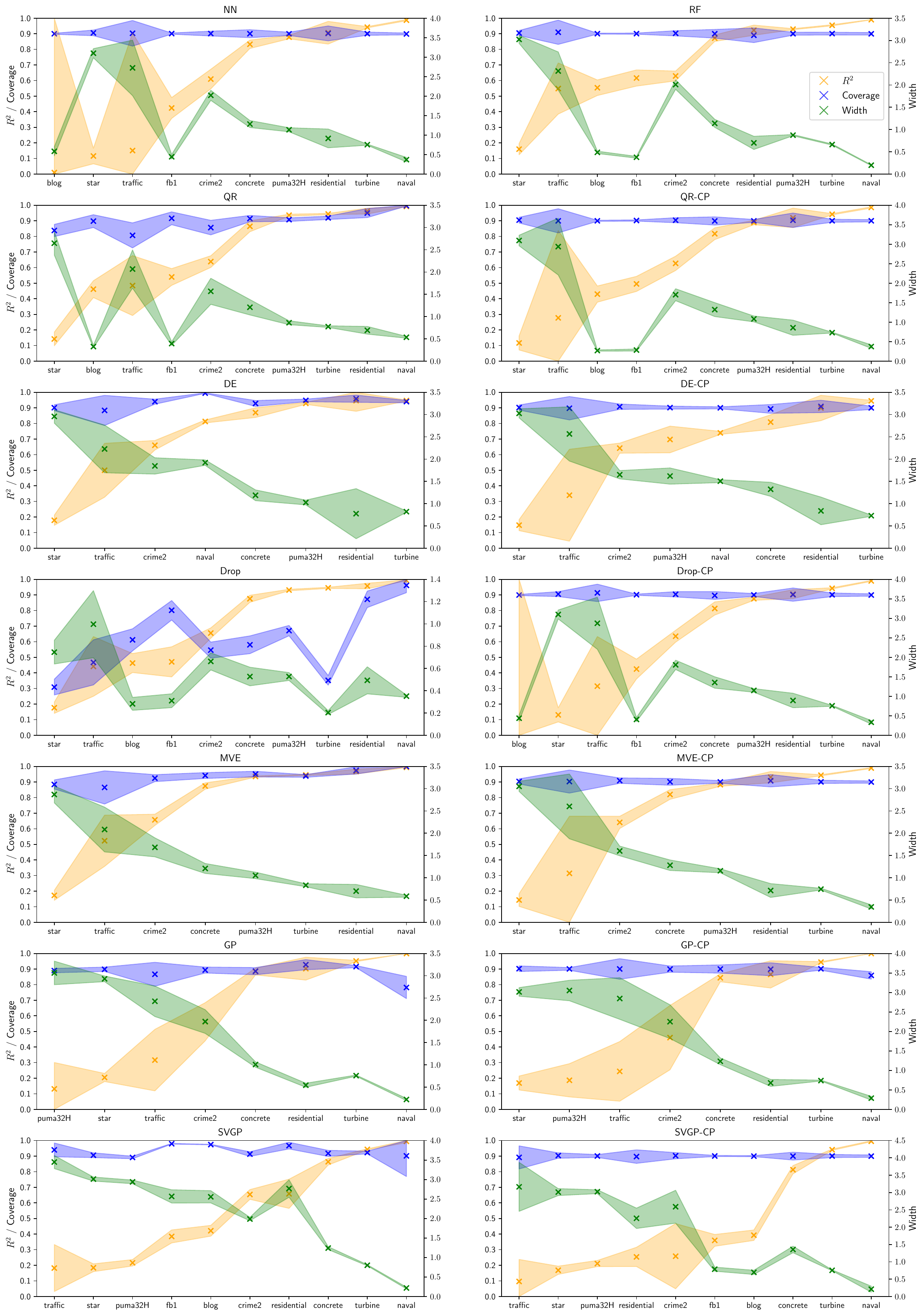}
        \caption{Relation between confidence and predictive power. For each model, the $R^2$-coefficient, coverage and average PI width are shown for all data sets in orange, blue and green, respectively. Shaded areas represent the standard deviation. For each model, the data sets are sorted along the x-axis according to the $R^2$-coefficient (in ascending order).}
        \label{fig:r2_comparison}
    \end{figure}

    In Fig.~\ref{fig:r2_comparison}, both the coverage degree, average width and $R^2$-coefficient are shown. For each model, the data sets are sorted according to increasing $R^2$-coefficient (averaged over the different runs). For the exact Gaussian processes (GP), no results on the data sets \texttt{fb1} and \texttt{blog} are reported due to the large memory and time consumption. While these sets are not extraordinary compared to modern real-life data sets, they already require a considerable amount of memory due to the quadratic scaling. Approximate models, such as the variational approximation used in this study, will become imperative. The first row of Fig.~\ref{fig:r2_comparison} consists of the two conformalized point predictors (these could be considered baseline models). For the other rows, the left column shows the results of the models trained on the full data set, while the right column shows those of the calibrated (conformalized) models. When comparing the two columns, it is immediately clear that the coverage, indicated by the blue regions, is much more concentrated around the nominal value of 0.9 for the conformalized models, as is guaranteed by the Marginal Validity Theorem from Section \ref{section:cp}. The shaded regions indicate the variability among the 50 different runs (standard deviation is used for Fig.~\ref{fig:r2_comparison}). It is clear that a higher variability in the predictive power often corresponds to a higher variability in the interval quality (both coverage and average width). This is not surprising since a model in general performs worse when the uncertainty is higher. As most of the models explicitly use the predictions to build prediction intervals, this relation can be expected to be even stronger. For both the DE and MVE models the results for the \texttt{fb1} and \texttt{blog} data sets are missing because the average widths differed by about two orders of magnitude compared to the other data sets and models and were, therefore, deemed to be nonsensical. The $R^2$-coefficient for the NN and Drop-CP models on the \texttt{blog} data set show a strong variability, while the average width remains small and almost constant. This can be explained by the strong skewness present in the data set. The $R^2$-coefficient is sensitive to the extreme outliers, while the prediction intervals are not, as long as the outlier proportion is less than $\alpha$. This also explains why almost all models give reasonably good intervals for both the \texttt{fb1} and \texttt{blog} data sets. The DE and MVE models form the exception, as mentioned above, since these methods inherently take into account the data uncertainty and cannot discard these outliers. A general summary of the results can be found in the tables in Appendix~\ref{section:app1}.

\begin{figure}[t!]
        \centering
        \includegraphics[width=\textwidth,keepaspectratio]{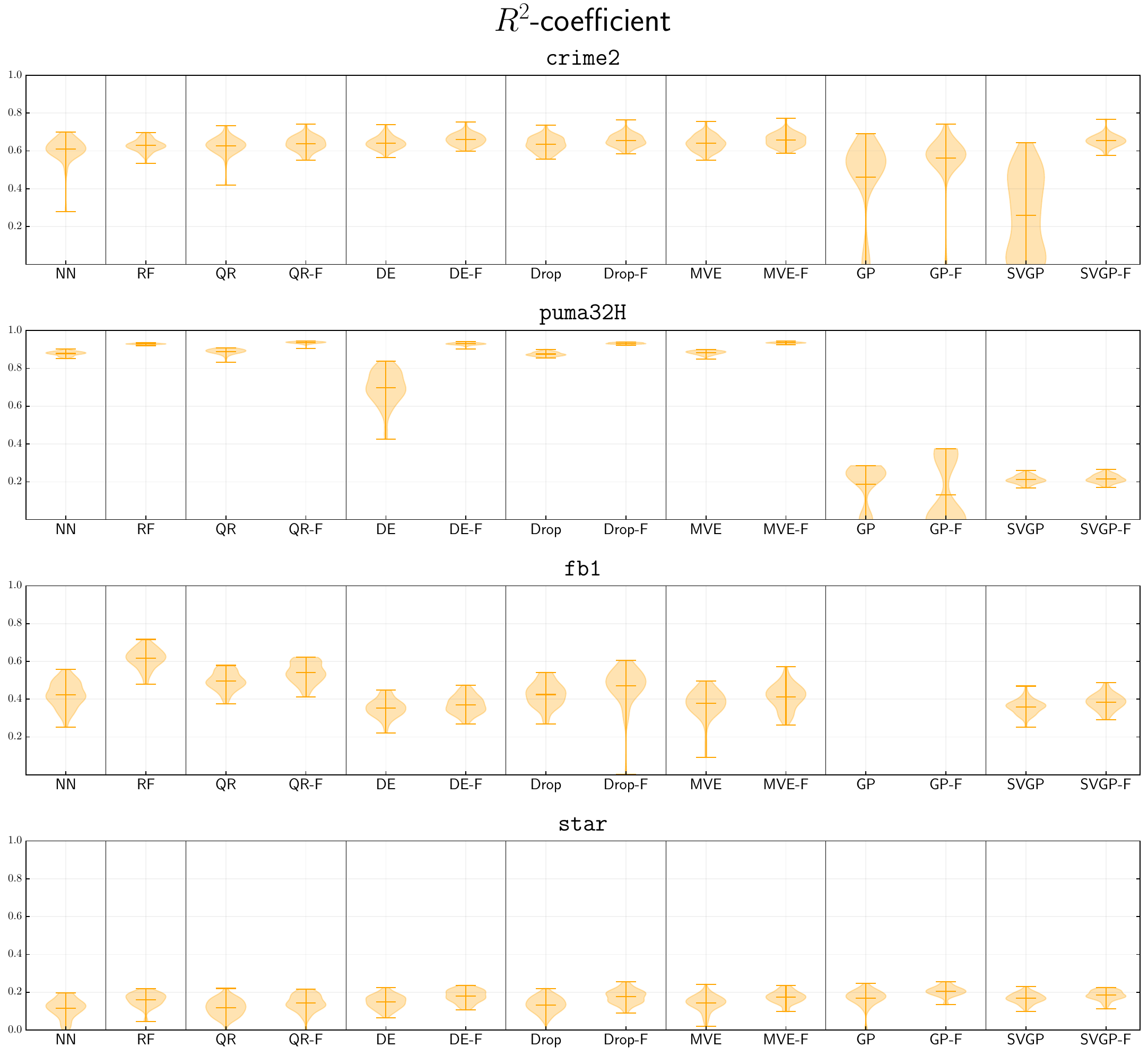}
        \caption{$R^2$-coefficients for a selection of data sets. For every model 2 variants are shown (except for NN and RF): trained on the full data set and on half of the data set (as for the conformalized models). The fully trained models are indicated by the tag `-F'.}
        \label{fig:R2_all}
    \end{figure}

    In Fig.~\ref{fig:PI_all} the coverage degree and the average interval width are shown for a selection of data sets (\texttt{crime2}, \texttt{turbine}, \texttt{fb1} and \texttt{star}). Except for NN and RF, every model is reported in three variants. The first one is trained on half of the data set, the second one is trained on all of the data set and the third one is conformalized using a 50/50 split. As before, the average widths for MVE and DE are not shown for the data set \texttt{fb1} because these were too large to be sensible. The associated $R^2$-coefficients are shown in Fig.~\ref{fig:R2_all}.
    \begin{figure}[t!]
        \centering
        \includegraphics[width=\textwidth,keepaspectratio]{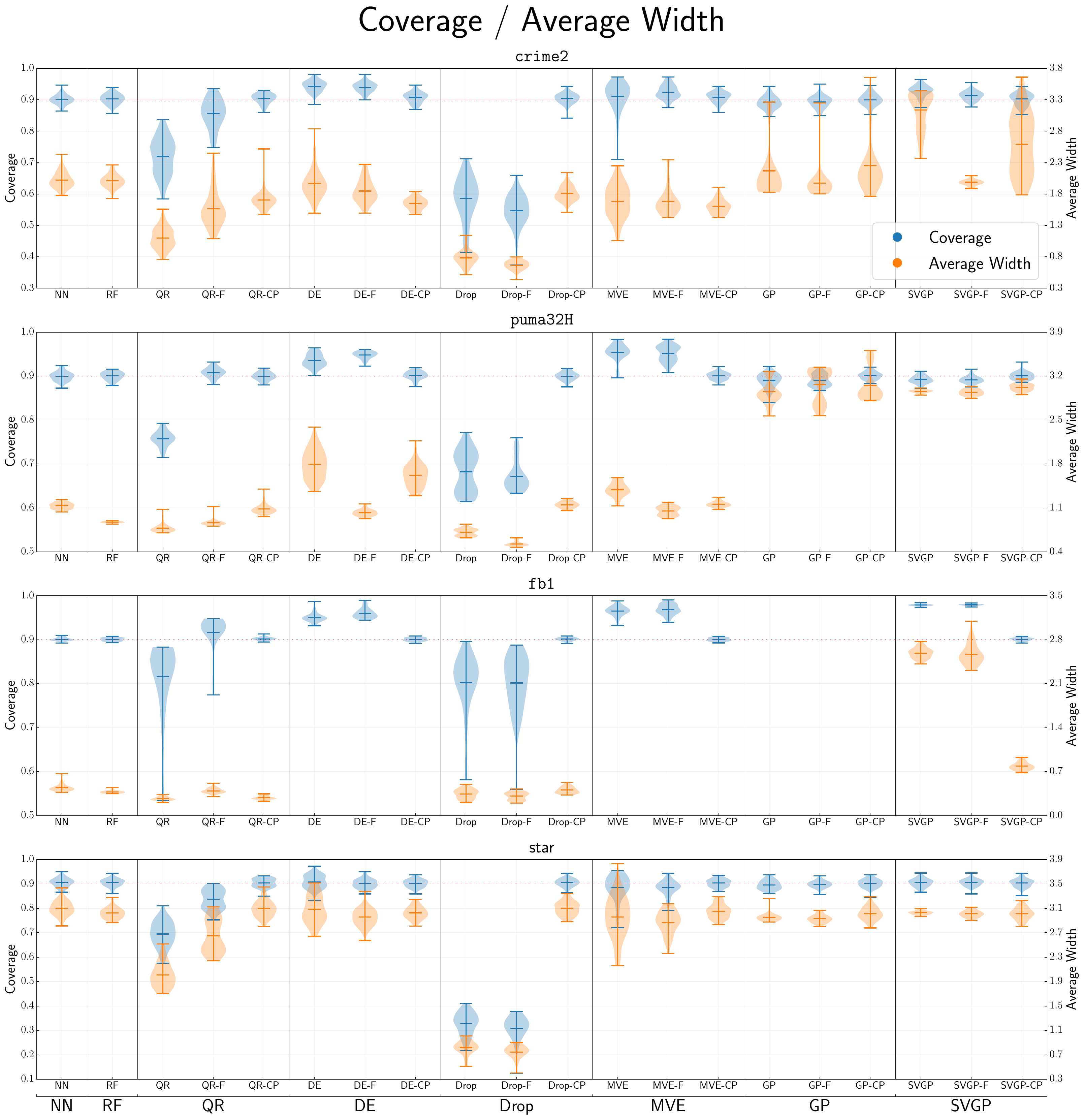}
        \caption{Prediction interval properties for a selection of data sets. Blue and orange regions respectively show the coverage degree and the average interval width. For every model 3 variants are shown (except for NN and RF). Fully trained and conformalized models are indicated by F and CP, respectively. The red dashed line indicates the nominal coverage level.}
        \label{fig:PI_all}
    \end{figure}
    The results for the conformalized models are the same as for the those trained on half of the data set, since conformal prediction is a post-hoc method. Therefore, only the fully-trained and conformalized models are shown. Here again it is clear that the uncalibrated models do not approximately saturate the validity constraint. They either underestimate the uncertainty or produce overconservative prediction intervals. When comparing between the models trained on half of the data set and the full data set, it is generally the case that the fully-trained model achieves better results in terms of both coverage and average width. The models incorporating a probabilistic prior (DE, MVE and GP) come out as the most stable ones w.r.t.\ data size change.

    \begin{figure}[p]
        \centering
        \includegraphics[width=\textwidth,keepaspectratio]{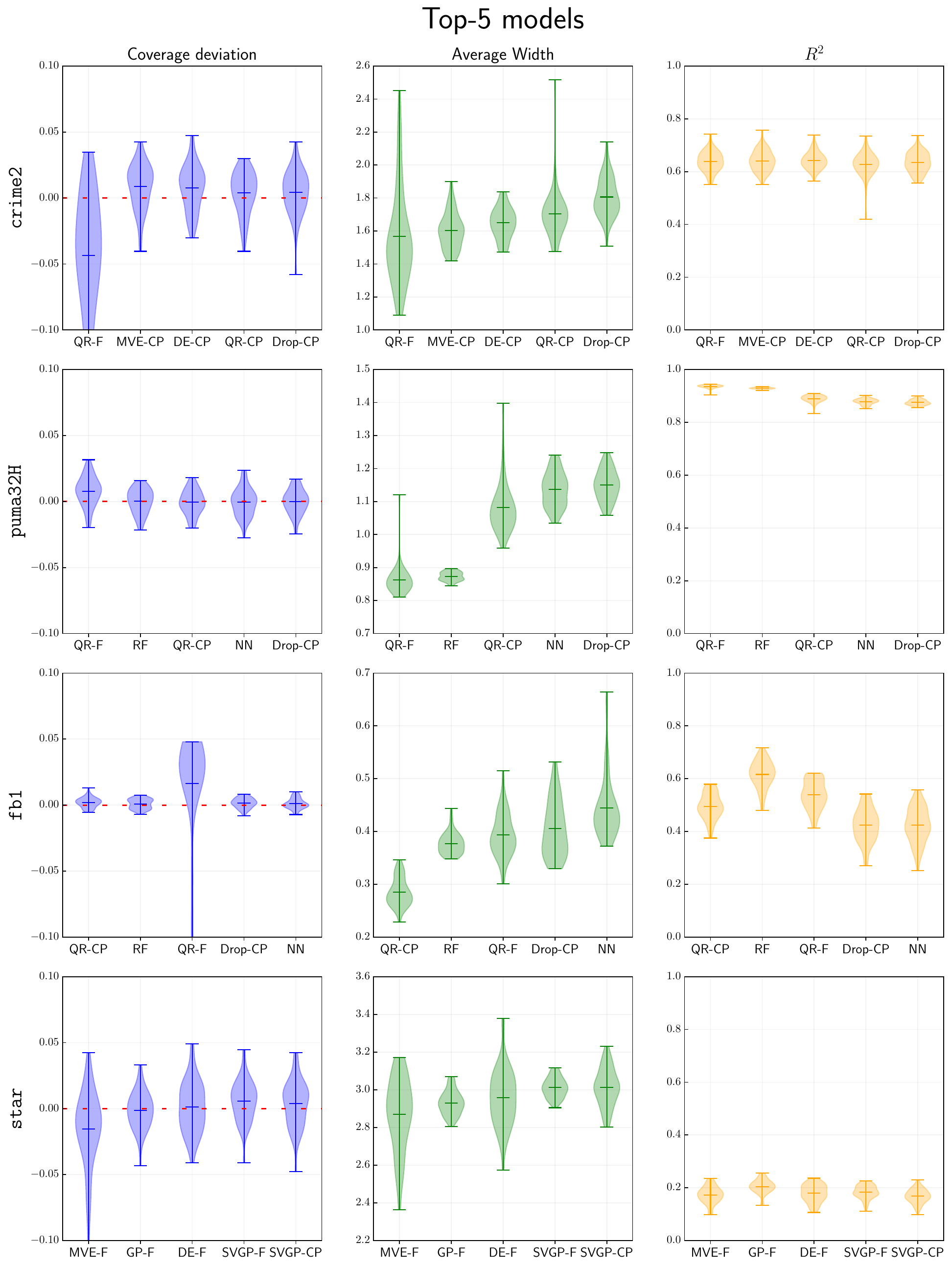}
        \caption{Overview of the best models based on average width. The left column shows the deviation of the coverage from the desired level ($\alpha = 0.9$), the middle column shows the average width of the prediction intervals and the right column shows the $R^2$-coefficient of the models. Error bars represent the standard deviation over train-test splits.}
        \label{fig:top5}
    \end{figure}

    For each of the selected models, Fig.~\ref{fig:top5} shows the best five models in terms of average width, excluding those that do not (approximately) satisfy the coverage constraint \eqref{calibrated}. This figure shows that there is quite some variation in the models. There is not a clear best choice. Because on most data sets the models produce uncalibrated prediction intervals, post-hoc calibrated models often produce the best intervals. However, when the models already produce (approximately) calibrated intervals from the start, they tend to give tighter intervals than their conformalized counterparts because there is more data available for training. It is also found that the models incorporating a Gaussian assumption (GP, SVGP, MVE and DE) fail to deliver optimal results in the case of non-Gaussian data (e.g.\ the \texttt{fb1} data). The desired confidence level can be attained, but there is a large trade-off with the average width. As noted before, the models (MVE and DE) using scoring rule~\eqref{gaussian_score} sometimes produce intervals where the average width is some orders of magnitude larger than for the other models. Leaving out the inherent uncertainty contribution to the variance (the third term in Eq.~\eqref{dropout_mve_var}) resolves this problem, but then the model loses its ability to approximately reach the required coverage level, similar to the ordinary dropout networks. However, as stated in the previous section, the data sets \texttt{fb1} and  \texttt{blog} are strongly skewed. After reducing the skewness by discarding data points greater than a certain threshold, the average width also strongly decreases. This observation, combined with the fact that it are exactly the models explicitly incorporating data (or \textit{aleatoric} \cite{kendallgal}) uncertainty that suffer from extreme prediction intervals, might also point to a different interpretation. Namely, in the case of outliers, the model has almost no information of the conditional distribution, hence this corresponds to a situation of almost full uncertainty. From this point of view, an extremely wide prediction interval might be more sensible than a mild extrapolation of the lower uncertainty on the majority of the data points.

    Because all data sets were standardized, the average widths of the prediction intervals can also be compared across data sets. From this perspective, the models seem to give significantly better results on the \texttt{fb1} data set. However, when making such a comparison, one should also look at the distribution of the data itself. The difference between the $(1-\alpha/2)$- and $\alpha/2$-quantiles gives a rough estimate of the interval widths. The values of this quantity for the four selected data sets, averaged over the train-test splits, are given by $(3.17, 3.49, 0.86, 3.24)$. Similar to the \textit{relative absolute error}, one could define a ``relative width'' as the ratio of the average width as estimated by the given model and the quantile difference obtained from the unconditional data distribution, i.e. the interval width one would obtain when ignoring the underlying features in the data set. If this quantity is smaller than one, the model is able to improve on a simple featureless estimator. From this point of view, it is clear that the models do not perform as well as could be thought on first sight for the \texttt{fb1} data set. At the same time it is also clear that for the \texttt{star} data set, the models only barely improve on the naive estimate.

    To see the influence of the training-calibration split on the resulting prediction intervals, two smaller experiments were performed where the training-calibration ratio was modified. In the first experiment the split ratio was changed from 50/50 to 75/25, i.e. more data was reserved for the training step. The average coverage was not significantly changed. However, the average width of the intervals decreased on average by 7\%. Using more data to train the underlying models, thereby obtaining better predictions, will lead to tighter prediction intervals as long as the calibration set is not too small. This conclusion is in line with the observations from Figs.~\ref{fig:r2_comparison} and \ref{fig:PI_all}. This experiment was then repeated in an extreme fashion, where the models were trained on the full data set. Due to the lack of an independent calibration set, the post-hoc calibration was also performed on the same training set. This way the influence of violating the assumptions of the ICP Algorithm~\ref{algorithm:icp} and the associated validity theorem can also be investigated. It was found that for some models the coverage decreased sharply. Both of these observations should not come as a surprise. The amount by which the intervals are scaled can be interpreted as a hyperparameter of the model. In general it is better to use more data to train than to validate, as long as the validation data set is representative of the true population. Moreover, optimizing hyperparameters on the training set is known to lead to overfitting, which in this case corresponds to overly optimistic prediction intervals. The CP algorithm looks at how large the errors are on the calibration set, so as to be able to correct for them in the future. However, by using the training set to calibrate, future errors are underestimated and, therefore, the CP algorithm cannot fully correct for them.

\section{Conclusions and future perspectives}

    In this study several types of prediction interval estimators for regression problems were reviewed and compared. Two main properties were taken into account: the coverage degree and the average width of the prediction intervals. It was found that without post-hoc calibration the methods derived from a probabilistic model attained the best coverage degree. However, after calibration, all methods attained the desired coverage and in certain cases the calibrated models even produced intervals with a smaller average width. It was also observed that the predictive power of the model and the quality of the prediction intervals are correlated. For the post-hoc calibration step the conformal prediction framework was used. To obtain the desired results, this method requires the data set to be split in a training and a calibration set. A small experiment confirmed that violating this condition leads to strongly uncalibrated intervals. Although this can be expected from an overfitting point of view, this also shows that the assumption does not merely serve a theoretical purpose. On the other hand, increasing the relative size of the training set can result in smaller prediction intervals without having a negative influence on the calibration.

    Although a variety of methods was considered, it is not feasible to include all of them. The most important omission is a more detailed overview of Bayesian neural networks (although one can argue, as was done in the section on dropout networks, that some common neural networks are, at least partially, Bayesian by nature). The main reason for this omission is the large number of choices in terms of priors and approximations, both of which strongly depend on the problem at hand. On the level of calibration there are also some methods that were not included in this paper, mostly because they were either too specific or too complex for simple regression problems. For general regression models the literature on calibration methods is not as extensive as it is for classification models. Recently some advances were made in which $\beta$-calibration \cite{pmlr-v54-kull17a} was generalized to regression problems using a Gaussian process approach~\cite{pmlr-v97-song19a}. However, as mentioned before, a Gaussian process does not have a favorable scaling behaviour and also in this case certain approximations are necessary. Another technique that was recently introduced~\cite{pmlr-v80-kuleshov18a} calibrates the cumulative distribution function produced by a distribution predictor using isotonic regression. Although the technique itself is simple in spirit, it is only applicable to predictors that construct the full output distribution. In a similar vein \cite{utpala2020quantile}, at the time of writing still under review, takes a distribution predictor and modifies the loss function such that the quantiles are calibrated without post-hoc calibration. The main benefit of this method is that it does not require a separate calibration set in stark contrast to conformal prediction, but it still requires the construction of the cumulative distribution. By dividing the target space in a finite number of bins, Keren et al.\ introduced an approach where the regression problem is approximated by a classification problem such that the usual tools for classifier calibration can be applied \cite{keren2018calibrated}. The main downside of this approach is that one loses the continuous nature of the initial problem. Another concept that was not covered is that of \textit{predictive distributions}~\cite{schweder2016confidence,shen2018prediction}, where not only a single interval is considered, but a full distribution is estimated. This approach was combined with conformal prediction in \cite{vovk2017nonparametric} giving rise to \textit{Conformal Predictive Systems}.

    The choice of data sets in this comparative study was very broad and no specific properties were taken into account a priori. After comparing the results of the different models, it did become apparent that certain assumptions or properties can have a major influence on the performance of the models. The main examples were the normality assumption for mean-variance estimators or the proper scoring rule~\eqref{gaussian_score}. Models using this scoring rule appeared to behave very badly when used for strongly skewed data sets. Since such data sets are gaining importance in the digital age, it would be interesting to both study methods tailored to these properties and how existing models behave on outliers. Another type of data that was not specifically considered in this study, but that is also becoming increasingly important, is time series data. The main issue with this kind of data, as was already mentioned in Section 4.2, is the autocorrelation, which (possibly) invalidates methods or theorems that make use of i.i.d.~or exchangeability properties. This has an influence on both the construction of models and the structure of validation sets \cite{chernozhukov2018exact}.

    A further aspect that was not considered in this study is the conditional behaviour of the models. When constructing a model that optimizes the coverage probability~\eqref{coverage}, only the \textit{marginal} coverage is controlled, i.e.~the specific properties of an instance are not taken into account. In certain cases it might be relevant to not only attain global validity, but also guarantee the validity on a certain subset of the instance space. A general notion of conditional validity was considered in \cite{lei2014distribution} and \cite{vovk2012conditional}. There it was also shown how to incorporate this notion in the conformal prediction framework.

\bibliographystyle{spmpsci}
\bibliography{Biblio}

\begin{appendices}
\section{Experimental results}\label{section:app1}

    In this section all experimental results are summarized. For every metric ($R^2$-coefficient, coverage and average width) the results are split into two tables, with five data sets each, for clarity. The rows represent the various models and the columns correspond to the different data sets. For every combination the average over 50 random train-test splits is reported. The standard deviation is shown between parentheses. As mentioned in Section~\ref{section:results}, there are no results available for exact Gaussian processes on the data sets \texttt{fb1} and \texttt{blog} due to exceeding the runtime limit; this is indicated by ``OoT'' (out of time). Moreover, certain values were deemed out-of-range, e.g.\ $R^2$-coefficients below -1 or cases where the standard deviation was unreasonably large; this is indicated by ``OoR'' (out of range).

    \begin{table}[ht!]
        \scriptsize
        \centering
        \renewcommand{\arraystretch}{1.5}
        \begin{tabular}{c||c|c|c|c|c}
            &\texttt{concrete}&\texttt{naval}&\texttt{turbine}&\texttt{puma32H}&\texttt{residential}\\
            \hhline{======}
            NN-CP&0.833 (0.027)&0.987 (0.004)&0.942 (0.004)&0.879 (0.011)&0.907 (0.073)\\
            \hdashline
            RF-CP&0.872 (0.021)&0.990 (0.002)&0.955 (0.004)&0.930 (0.003)&0.923 (0.033)\\
            \hdashline
            QR&0.865 (0.035)&0.992 (0.002)&0.944 (0.004)&0.936 (0.006)&0.962 (0.018)\\
            QR-CP&0.817 (0.038)&0.986 (0.004)&0.942 (0.004)&0.889 (0.014)&0.919 (0.062)\\
            \hdashline
            DE&0.870 (0.031)&0.813 (0.009)&0.947 (0.004)&0.929 (0.007)&0.946 (0.068)\\
            DE-CP&0.809 (0.047)&0.740 (0.010)&0.945 (0.004)&0.697 (0.085)&0.899 (0.081)\\
            \hdashline
            Drop&0.875 (0.023)&0.994 (0.003)&0.945 (0.004)&0.931 (0.005)&0.957 (0.018)\\
            Drop-CP&0.813 (0.042)&0.990 (0.003)&0.943 (0.004)&0.876 (0.010)&0.905 (0.029)\\
            \hdashline
            MVE&0.874 (0.020)&0.994 (0.002)&0.945 (0.004)&0.935 (0.004)&0.967 (0.016)\\
            MVE-CP&0.820 (0.031)&0.990 (0.003)&0.943 (0.004)&0.883 (0.010)&0.931 (0.036)\\
            \hdashline
            GP&0.884 (0.022)&1.000 (0.000)&0.952 (0.004)&0.132 (0.169)&0.903 (0.073)\\
            GP-CP&0.844 (0.025)&1.000 (0.000)&0.945 (0.004)&0.187 (0.107)&0.866 (0.087)\\
            \hdashline
            SVGP&0.864 (0.021)&0.995 (0.008)&0.943 (0.004)&0.216 (0.021)&0.659 (0.094)\\
            SVGP-CP&0.813 (0.027)&0.995 (0.003)&0.941 (0.004)&0.211 (0.020)&0.254 (0.062)\\
        \end{tabular}
        \caption{Summary of $R^2$-coefficients (part I). For the data sets in each column, the results are shown for all models (standard deviation between parentheses). For clarity, models that come in both a default and a conformalized version have been grouped together.}
        \label{table:r2coeff}
    \end{table}

    \begin{table}[ht!]
        \scriptsize
        \centering
        \renewcommand{\arraystretch}{1.5}
        \begin{tabular}{c||c|c|c|c|c}
            &\texttt{crime2}&\texttt{fb1}&\texttt{blog}&\texttt{traffic}&\texttt{star}\\
            \hhline{======}
            NN-CP&0.609 (0.063)&0.424 (0.067)&0.010 (1.390)&0.151 (0.755)&0.116 (0.050)\\
            \hdashline
            RF-CP&0.629 (0.032)&0.616 (0.052)&0.554 (0.050)&0.549 (0.164)&0.160 (0.036)\\
            \hdashline
            QR&0.638 (0.037)&0.540 (0.054)&0.462 (0.054)&0.485 (0.192)&0.142 (0.044)\\
            QR-CP&0.627 (0.046)&0.495 (0.048)&0.430 (0.052)&0.277 (0.560)&0.117 (0.045)\\
            \hdashline
            DE&0.660 (0.031)&0.371 (0.047)&0.289 (0.034)&0.500 (0.173)&0.179 (0.033)\\
            DE-CP&0.642 (0.032)&0.353 (0.048)&0.287 (0.031)&0.340 (0.295)&0.148 (0.035)\\
            \hdashline
            Drop&0.656 (0.033)&0.471 (0.097)&0.463 (0.061)&0.441 (0.191)&0.177 (0.036)\\
            Drop-CP&0.635 (0.038)&0.425 (0.064)&OoR&0.315 (0.318)&0.131 (0.045)\\
            \hdashline
            MVE&0.658 (0.036)&0.411 (0.065)&OoR&0.523 (0.164)&0.173 (0.032)\\
            MVE-CP&0.641 (0.039)&0.379 (0.074)&OoR&0.314 (0.366)&0.143 (0.041)\\
            \hdashline
            GP&0.562 (0.122)&OoT&OoT&0.316 (0.197)&0.205 (0.026)\\
            GP-CP&0.460 (0.206)&OoT&OoT&0.244 (0.191)&0.169 (0.045)\\
            \hdashline
            SVGP&0.654 (0.031)&0.385 (0.042)&0.421 (0.035)&0.182 (0.151)&0.184 (0.025)\\
            SVGP-CP&0.258 (0.209)&0.360 (0.038)&0.393 (0.034)&0.096 (0.141)&0.168 (0.026)
        \end{tabular}
        \caption{Summary of $R^2$-coefficients (part II). For the data sets in each column, the results are shown for all models (standard deviation between parentheses). For clarity, models that come in both a default and a conformalized version have been grouped together. ``OoT'' stands for ``out of time'', for these combinations of data sets and models, the maximum runtime was exceeded. ``OoR'' stands for ``out of range'', these values exceeded reasonable ranges for the $R^2$-coefficient.}
        \label{table:r2coeff2}
    \end{table}

    \begin{table}[ht!]
        \scriptsize
        \centering
        \renewcommand{\arraystretch}{1.5}
        \begin{tabular}{c||c|c|c|c|c}
            &\texttt{concrete}&\texttt{naval}&\texttt{turbine}&\texttt{puma32H}&\texttt{residential}\\
            \hhline{======}
            NN-CP&0.901 (0.025)&0.899 (0.006)&0.900 (0.010)&0.900 (0.011)&0.902 (0.048)\\
            \hdashline
            RF-CP&0.900 (0.028)&0.900 (0.008)&0.901 (0.009)&0.900 (0.009)&0.890 (0.047)\\
            \hdashline
            QR&0.914 (0.020)&0.997 (0.003)&0.919 (0.011)&0.908 (0.011)&0.950 (0.029)\\
            QR-CP&0.899 (0.026)&0.900 (0.007)&0.900 (0.010)&0.899 (0.009)&0.904 (0.046)\\
            \hdashline
            DE&0.928 (0.019)&0.995 (0.002)&0.940 (0.008)&0.948 (0.008)&0.959 (0.023)\\
            DE-CP&0.893 (0.028)&0.900 (0.007)&0.900 (0.011)&0.902 (0.010)&0.909 (0.040)\\
            \hdashline
            Drop&0.580 (0.057)&0.962 (0.047)&0.352 (0.031)&0.671 (0.033)&0.872 (0.053)\\
            Drop-CP&0.897 (0.025)&0.900 (0.007)&0.901 (0.012)&0.900 (0.009)&0.902 (0.043)\\
            \hdashline
            MVE&0.942 (0.019)&1.000 (0.001)&0.939 (0.009)&0.951 (0.018)&0.974 (0.023)\\
            MVE-CP&0.901 (0.022)&0.900 (0.007)&0.901 (0.010)&0.901 (0.008)&0.908 (0.040)\\
            \hdashline
            GP&0.887 (0.023)&0.782 (0.071)&0.916 (0.007)&0.890 (0.014)&0.928 (0.032)\\
            GP-CP&0.900 (0.026)&0.859 (0.024)&0.902 (0.009)&0.901 (0.009)&0.899 (0.041)\\
            \hdashline
            SVGP&0.918 (0.020)&0.902 (0.132)&0.923 (0.009)&0.891 (0.008)&0.967 (0.022)\\
            SVGP-CP&0.900 (0.025)&0.900 (0.008)&0.901 (0.010)&0.901 (0.009)&0.897 (0.043)\\
        \end{tabular}
        \caption{Summary of coverage degrees (part I). For the data sets in each column, the results are shown for all models (standard deviation between parentheses). For clarity, models that come in both a default and a conformalized version have been grouped together.}
        \label{table:coverage}
    \end{table}

    \begin{table}[ht!]
        \scriptsize
        \centering
        \renewcommand{\arraystretch}{1.5}
        \begin{tabular}{c||c|c|c|c|c}
            &\texttt{crime2}&\texttt{fb1}&\texttt{blog}&\texttt{traffic}&\texttt{star}\\
            \hhline{======}
            NN-CP&0.901 (0.015)&0.901 (0.004)&0.900 (0.003)&0.903 (0.082)&0.906 (0.018)\\
            \hdashline
            RF-CP&0.903 (0.017)&0.901 (0.004)&0.901 (0.003)&0.910 (0.078)&0.905 (0.016)\\
            \hdashline
            QR&0.857 (0.047)&0.916 (0.041)&0.898 (0.041)&0.807 (0.080)&0.838 (0.039)\\
            QR-CP&0.904 (0.016)&0.902 (0.004)&0.900 (0.004)&0.900 (0.077)&0.904 (0.018)\\
            \hdashline
            DE&0.940 (0.016)&0.960 (0.011)&0.959 (0.010)&0.884 (0.097)&0.901 (0.019)\\
            DE-CP&0.908 (0.017)&0.901 (0.004)&0.900 (0.003)&0.898 (0.075)&0.902 (0.016)\\
            \hdashline
            Drop&0.546 (0.051)&0.802 (0.062)&0.612 (0.071)&0.467 (0.145)&0.309 (0.050)\\
            Drop-CP&0.904 (0.017)&0.901 (0.004)&0.900 (0.004)&0.913 (0.056)&0.905 (0.016)\\
            \hdashline
            MVE&0.925 (0.022)&0.969 (0.012)&0.965 (0.012)&0.865 (0.107)&0.884 (0.029)\\
            MVE-CP&0.909 (0.017)&0.901 (0.004)&0.900 (0.003)&0.903 (0.074)&0.904 (0.017)\\
            \hdashline
            GP&0.894 (0.019)&OoT&OoT&0.867 (0.077)&0.899 (0.014)\\
            GP-CP&0.900 (0.020)&OoT&OoT&0.901 (0.066)&0.902 (0.018)\\
            \hdashline
            SVGP&0.913 (0.015)&0.979 (0.002)&0.974 (0.002)&0.940 (0.044)&0.906 (0.015)\\
            SVGP-CP&0.903 (0.020)&0.901 (0.004)&0.900 (0.003)&0.892 (0.074)&0.904 (0.017)\\
        \end{tabular}
        \caption{Summary of coverage degrees (part II). For the data sets in each column, the results are shown for all models (standard deviation between parentheses). For clarity, models that come in both a default and a conformalized version have been grouped together. ``OoT'' stands for ``out of time'', for these combinations of data sets and models, the maximum runtime was exceeded.}
        \label{table:coverage2}
    \end{table}

    \begin{table}[ht!]
        \scriptsize
        \centering
        \renewcommand{\arraystretch}{1.5}
        \begin{tabular}{c||c|c|c|c|c}
            &\texttt{concrete}&\texttt{naval}&\texttt{turbine}&\texttt{puma32H}&\texttt{residential}\\
            \hhline{======}
            NN-CP&1.284 (0.090)&0.370 (0.057)&0.754 (0.014)&1.136 (0.052)&0.915 (0.241)\\
            \hdashline
            RF-CP&1.139 (0.090)&0.199 (0.010)&0.663 (0.013)&0.873 (0.014)&0.700 (0.148)\\
            \hdashline
            QR&1.209 (0.173)&0.535 (0.021)&0.775 (0.018)&0.862 (0.044)&0.693 (0.087)\\
            QR-CP&1.322 (0.179)&0.374 (0.046)&0.731 (0.011)&1.083 (0.075)&0.857 (0.195)\\
            \hdashline
            DE&1.188 (0.121)&1.921 (0.061)&0.820 (0.015)&1.026 (0.053)&0.775 (0.562)\\
            DE-CP&1.322 (0.158)&1.505 (0.033)&0.730 (0.015)&1.620 (0.182)&0.838 (0.311)\\
            \hdashline
            Drop&0.527 (0.084)&0.351 (0.010)&0.204 (0.017)&0.527 (0.035)&0.493 (0.122)\\
            Drop-CP&1.354 (0.147)&0.334 (0.047)&0.755 (0.012)&1.150 (0.047)&0.893 (0.186)\\
            \hdashline
            MVE&1.209 (0.112)&0.586 (0.022)&0.832 (0.033)&1.053 (0.072)&0.700 (0.151)\\
            MVE-CP&1.282 (0.118)&0.346 (0.047)&0.744 (0.020)&1.157 (0.044)&0.714 (0.155)\\
            \hdashline
            GP&1.006 (0.060)&0.222 (0.025)&0.757 (0.013)&3.064 (0.265)&0.542 (0.046)\\
            GP-CP&1.234 (0.087)&0.287 (0.049)&0.739 (0.012)&3.050 (0.263)&0.679 (0.090)\\
            \hdashline
            SVGP&1.240 (0.038)&0.217 (0.021)&0.800 (0.015)&2.939 (0.047)&2.769 (0.224)\\
            SVGP-CP&1.356 (0.094)&0.210 (0.087)&0.754 (0.013)&3.023 (0.056)&2.256 (0.294)\\
        \end{tabular}
        \caption{Summary of average widths (part I). For the data sets in each column, the results are shown for all models (standard deviation between parentheses). For clarity, models that come in both a default and a conformalized version have been grouped together.}
        \label{table:width}
    \end{table}

    \begin{table}[ht!]
        \scriptsize
        \centering
        \renewcommand{\arraystretch}{1.5}
        \begin{tabular}{c||c|c|c|c|c}
            &\texttt{crime2}&\texttt{fb1}&\texttt{blog}&\texttt{traffic}&\texttt{star}\\
            \hhline{======}
            NN-CP&2.021 (0.130)&0.444 (0.059)&0.583 (0.118)&2.724 (0.711)&3.101 (0.117)\\
            \hdashline
            RF-CP&2.009 (0.106)&0.377 (0.020)&0.487 (0.027)&2.314 (0.428)&3.023 (0.090)\\
            \hdashline
            QR&1.567 (0.293)&0.394 (0.041)&0.327 (0.027)&2.067 (0.430)&2.648 (0.267)\\
            QR-CP&1.704 (0.153)&0.285 (0.026)&0.271 (0.016)&2.939 (0.729)&3.095 (0.136)\\
            \hdashline
            DE&1.847 (0.183)&OoR&OoR&2.229 (0.535)&2.958 (0.150)\\
            DE-CP&1.649 (0.093)&OoR&OoR&2.565 (0.609)&3.027 (0.102)\\
            \hdashline
            Drop&0.663 (0.077)&0.310 (0.062)&0.282 (0.058)&0.997 (0.301)&0.746 (0.106)\\
            Drop-CP&1.807 (0.120)&0.405 (0.053)&0.438 (0.081)&2.873 (0.668)&3.100 (0.115)\\
            \hdashline
            MVE&1.683 (0.213)&OoR&OoR&2.083 (0.503)&2.870 (0.189)\\
            MVE-CP&1.602 (0.109)&OoR&OoR&2.602 (0.727)&3.052 (0.114)\\
            \hdashline
            GP&1.972 (0.266)&OoT&OoT&2.427 (0.343)&2.931 (0.063)\\
            GP-CP&2.251 (0.433)&OoT&OoT&2.845 (0.534)&3.014 (0.111)\\
            \hdashline
            SVGP&1.986 (0.045)&2.567 (0.170)&2.556 (0.158)&3.449 (0.168)&3.012 (0.052)\\
            SVGP-CP&2.590 (0.475)&0.787 (0.059)&0.695 (0.061)&3.164 (0.707)&3.012 (0.100)\\
        \end{tabular}
        \caption{Summary of average widths (part II). For the data sets in each column, the results are shown for all models (standard deviation between parentheses). For clarity, models that come in both a default and a conformalized version have been grouped together.``OoT'' stands for ``out of time'', for these combinations of data sets and models, the maximum runtime was exceeded. ``OoR'' stands for ``out of range'', these values exceeded reasonable ranges for the average interval width.}
        \label{table:width2}
    \end{table}

\end{appendices}

\end{document}